\documentclass[journal=jacsat,manuscript=article]{achemso}
\usepackage[utf8]{inputenc}
\usepackage{amsmath}
\usepackage{amssymb}
\usepackage{graphicx}
\usepackage{algorithm}
\usepackage{algpseudocode}
\usepackage{subcaption}
\usepackage{arydshln}
\usepackage{color}
\usepackage{fixltx2e}
\usepackage{hyperref}

\usepackage{soul}
\usepackage[dvipsnames]{xcolor}

\usepackage[version=3]{mhchem} 

\author{Cooper Lorsung}
\affiliation[meche]
{Department of Mechanical Engineering, Carnegie Mellon University, USA}
\author{Zijie Li}
\affiliation[meche]
{Department of Mechanical Engineering, Carnegie Mellon University, USA}
\author{Amir Barati Farimani}
\email{barati@cmu.edu}
\affiliation[meche]
{Department of Mechanical Engineering, Carnegie Mellon University, USA}
\alsoaffiliation[biomed]
{Department of Biomedical Engineering, Carnegie Mellon University, USA}
\alsoaffiliation[mld]
{Machine Learning Department, Carnegie Mellon University, USA}

\title
  {Physics Informed Token Transformer for Solving Partial Differential Equations}

\begin{document}
\begin{abstract}
    Solving Partial Differential Equations (PDEs) is the core of many fields of science and engineering.
    While classical approaches are often prohibitively slow, machine learning models often fail to incorporate complete system information.
    Over the past few years, transformers have had a significant impact on the field of Artificial Intelligence and have seen increased usage in PDE applications.
    However, despite their success, transformers currently lack integration with physics and reasoning.
    This study aims to address this issue by introducing PITT: Physics Informed Token Transformer.
    The purpose of PITT is to incorporate the knowledge of physics by embedding partial differential equations (PDEs) into the learning process.
    PITT uses an equation tokenization method to learn an analytically-driven numerical update operator.
    By tokenizing PDEs and embedding partial derivatives, the transformer models become aware of the underlying knowledge behind physical processes.
    To demonstrate this, PITT is tested on challenging 1D and 2D PDE operator learning tasks.
    The results show that PITT outperforms popular neural operator models and has the ability to extract physically relevant information from governing equations.
\end{abstract}
\section*{Introduction}
\vspace{-4mm}
Partial Differential Equations (PDEs) are ubiquitous in science and engineering applications.
While much progress has been made in developing analytical and computational methods to solve the various equations, no complete analytical theory exists, and computational methods are often prohibitively expensive.
Recent work has shown the ability to learn analytical solutions using bilinear residual networks\cite{Zhang2022-dh}, and bilinear neural networks\cite{Zhang2019-yy, Zhang2021-rn, Zhang2021-ko}, where an analytical solution is available.
Many machine learning approaches have been proposed to improve simulation speed to calculate various fluid properties, where no such analytical solution is known to exist, including discrete mesh optimization\cite{meshdqn, pmlr-v206-yang23e, foucart2022deep, wu2023learning}, super resolution on lower resolution simulations\cite{fluid_superresolution, volumetric_gan}, and surrogate modeling\cite{dl_rom,hemmasian_surrogate_2023, li2021fourier, DeepONet-Nature-2021}.
While mesh optimization generally allows for using traditional numerical solvers, current methods only improve speed or accuracy by a few percent, or require many simulations during training.
Methods for super resolution improve speed, but often struggle with generalizing to data resolutions not seen in the training data, with more recent work improving generalization capabilities\cite{shu_physics-informed_2023}.
Surrogate modeling, on the other hand, has shown a good balance between improved performance and generalization.
Neural operator learning architectures, specifically, have also shown promise in combining super resolution capability with surrogate modeling due to their inherent discretization invariance.\cite{kovachki_neural_nodate}.
Recently, the attention mechanism has become a popular choice for operator learning.

The attention mechanism first emerged as a promising model for natural language processing tasks \cite{bahdanau2016nmt, vaswani_attention_2017, devlin2019bert, brown2020language}, especially the scaled dot-product attention \cite{vaswani_attention_2017}.
Its success has been extended to other areas, including computer vision tasks \cite{dosovitskiy2021image} and biology \cite{jumper2021highly}.
It has also inspired a wide array of scientific applications, in particular PDEs modeling \cite{cao_choose_2021, li2022transformer, hao2023gnot, temporal-attention-mesh-reduced-space-iclr2022, ovadia2023vito, Geneva_2022, guo2023transformer,romer}.
Kovachki \textit{et al.}\cite{kovachki2022neural} proposes a kernel integral interpretation of attention.
Cao \cite{cao_choose_2021} analyzes the theoretical properties of softmax-free dot product attention (also known as linear attention) and further proposes two interpretations of attention, such that it can be viewed as the numerical quadrature of a kernel integral operator or a Peterov-Galerkin projection.
OFormer (Operator Transformer) \cite{li2022transformer} extends the kernel integral formulation of linear attention by adding relative positional encoding\cite{su2022roformer} and using cross attention to flexibly handle discretization, and further proposes a latent marching architecture for solving forward time-dependent problems.
\citet{guo2023transformer} introduces attention as an instance-based learnable kernel for direct sampling method and demonstrates superiority on boundary value inverse problems.
LOCA (Learning Operators with Coupled Attention)\cite{kissas2022learning} uses attention weights to learn correlations in the output domain and enables sample-efficient training of the model.
GNOT (General Neural Operator Transformer for Operator Learning) \cite{hao2023gnot} proposes a heterogeneous attention architecture that stacks multiple cross-attention layers and uses a geometric gating mechanism to adaptively aggregate features from query points.
Additionally, encoding physics-informed inductive biases has also been of great interest because it allows incorporatation of additional system knowledge, making the learning task easier.
One strategy to encode the parameters of different instances for parametric PDEs is by adding conditioning module to the model \cite{wang2022metalearning, takamoto2023cape}. 
 Another approach is to embed governing equations into the loss function, known as Physics-Informed Neural Networks (PINNs)\cite{pinn}.
Physics Informed Neural Networks (PINNs) have shown promise in physics-based tasks, but have some downsides.
Namely, they show lack of generalization, and are difficult to train.
Complex training strategies have been developed in order to account for these deficiencies \cite{krishnapriyan2021characterizing}.

\begin{figure}
    \centering
    \begin{subfigure}[b]{0.8\linewidth}
        \includegraphics[width=\linewidth]{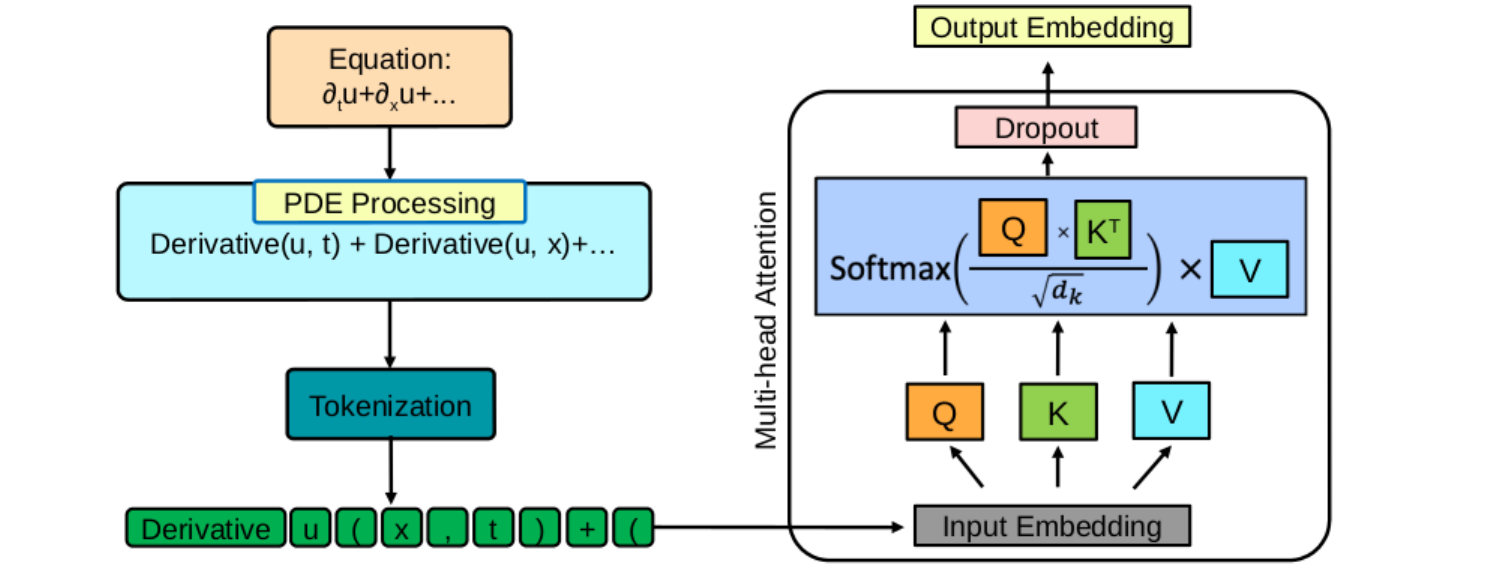}
        \caption{Token Transformer module.}
        \label{fig:token_transformer}
    \end{subfigure}
    \begin{subfigure}[b]{0.8\linewidth}
        \includegraphics[width=\linewidth]{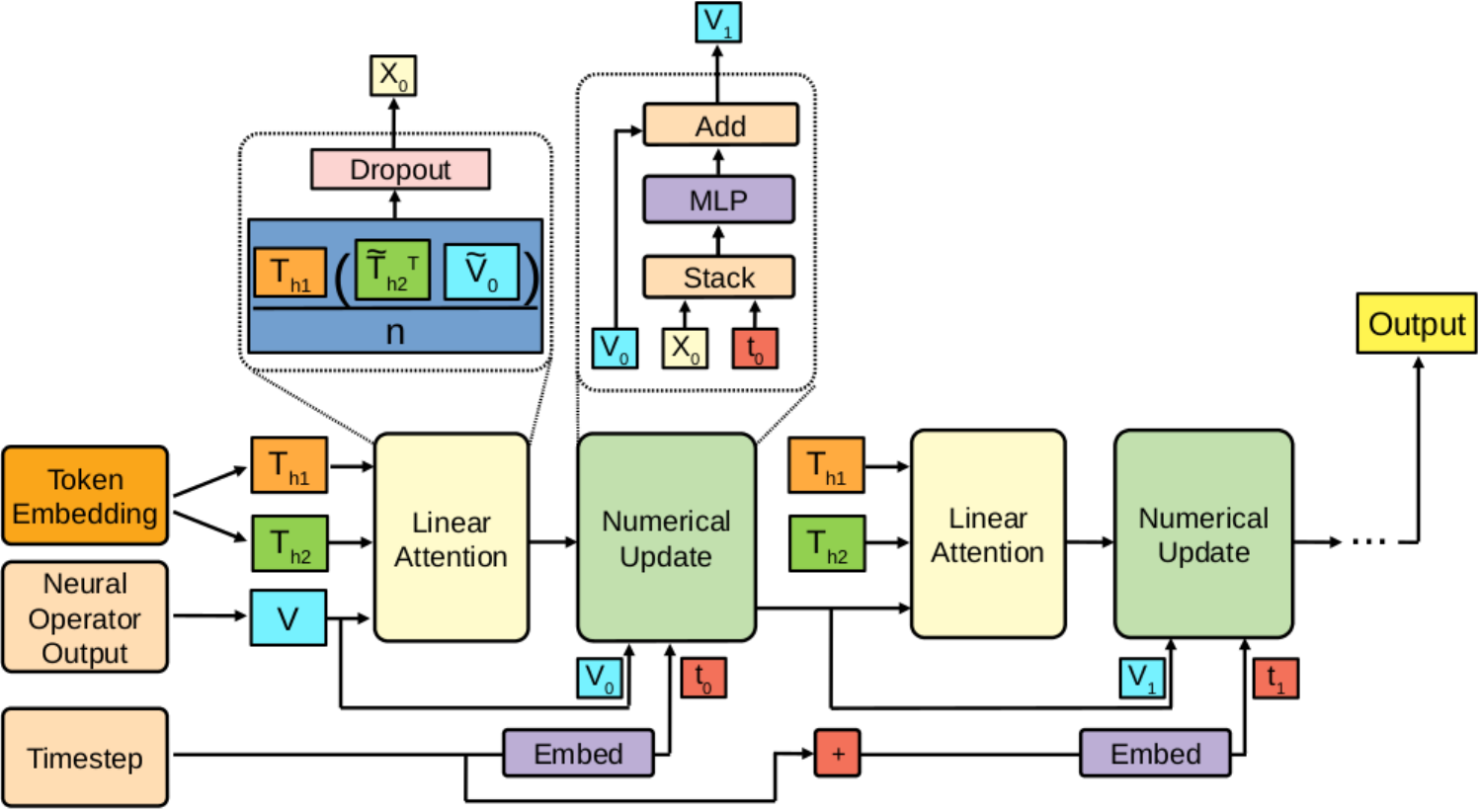}
        \caption{Linear Attention Update module.}
        \label{fig:linatt_update}
    \end{subfigure}
    \begin{subfigure}[b]{0.8\linewidth}
        \includegraphics[width=0.9\linewidth]{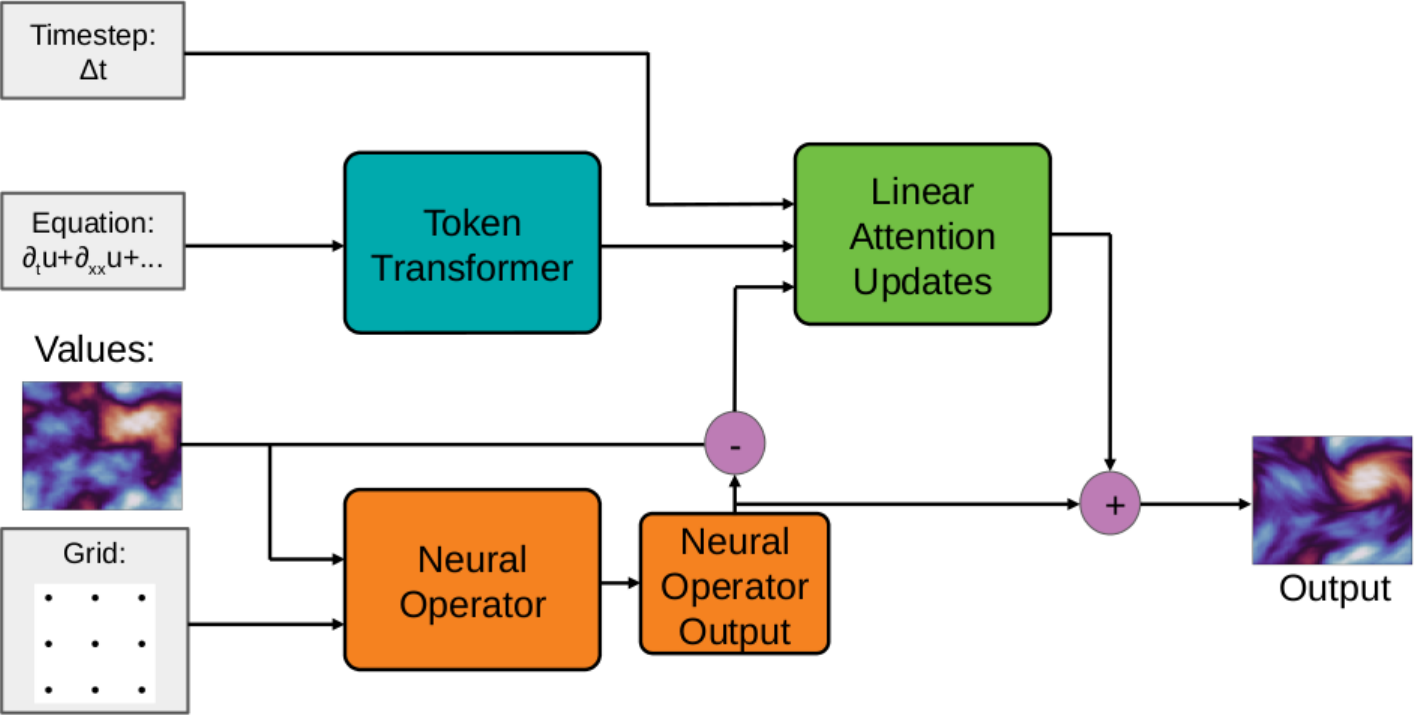}
        \caption{Physics Informed Token Transformer}
        \label{fig:PITT_framework}
    \end{subfigure}
    \caption{The Physics Informed Token Transformer (PITT) uses standard multi head self-attention to learn a latent embedding of the governing equations. This latent embedding is then used to perform numerical updates using linear attention blocks. The equation embedding acts as an analytically-driven correction to an underlying data-driven neural operator.}
    \label{fig:pitt}
\end{figure}

While many existing works are successful in their own right, none so far have incorporated entire analytical governing equations.
In this work we introduce an equation embedding strategy as well as an attention-based architecture, Physics Informed Token Transformer (PITT), to perform neural operator learning using equation information that utilizes physics-based inductive bias directly from governing equations (The main architecture of PITT is shown in figure \ref{fig:pitt}). More specifically, PITT fuses the equation knowledge into the neural operator learning by introducing a symbolic transformer on top of the neural operator.
We demonstrate through a series of challenging benchmarks that PITT outperforms the popular Fourier Neural Operator\cite{li2021fourier} (FNO), DeepONet\cite{DeepONet-Nature-2021}, and OFormer\cite{li2022transformer} and is able to learn physically relevant information from only the governing equations and system specifications.

\vspace{-4mm}
\vspace{-4mm}
\section{Methods}
\vspace{-4mm}
In this work, we aim to learn the operator $\mathcal{G}_{\theta}: \mathcal{A} \to \mathcal{U}$, where $\mathcal{A}$ is our input function space, $\mathcal{U}$ is our solution function space, and $\theta$ are the learnable model parameters.
We use a combination of novel equation tokenization and numerical method-like updates to learn model operators $\mathcal{G}_{\theta}$.
Our novel equation tokenization and embedding method is described first, followed by a detailed explanation of the numerical update scheme.

\vspace{-4mm}
\subsection{Equation Tokenization}
\vspace{-4mm}
\label{sec:eq_tokenization}
In order to utilize the text view of our data, the equations must be tokenized as input to our transformer.
Following Lampe \textit{et al.}\cite{lample_deep_2020}, each equation is parsed and split into its constituent symbols.
The tokens are given in table \ref{tab:tokens}.

\begin{table}[h]
    \begin{tabular}{c|c}
        \centering
         Category & Available Tokens \\
         \hline
         Equation & $($, $)$, $\partial$, $\Sigma$, $j$, $A_j$, $l_j$, $\omega_j$, $\phi_j$, $\sin$, $t$, $u$, $x$, $y$, $+$, $-$, $*$, $/$ \\
         Boundary Conditions & $Neumann, Dirichlet, None$ \\
         Numerical & $0$, $1, 2, 3, 4, 5, 6, 7, 8, 9, 10\string^, E$, $e$ \\
         Delimiter & , (comma), . (decimal point) \\
         Separator & \& \\
         2D Equations & $\nabla, =, \Delta, \cdot \text{ (dot product)}$
    \end{tabular}
    \caption{Collection of all tokens used in tokenizing governing equations, sampled values, and system parameters.}
    \label{tab:tokens}
\end{table}

%
%
The delimiter marks - decimal points for numerical values, commas for separating numerical values, and ampersand for separating equations, sampled values, and boundary conditions - are also added. 
The 1D equations are tokenized so the governing equation, forcing term, initial condition, sampled values, and output simulation time are all separated because each component controls distinct properties of the system.
The 2D equations are tokenized so that the governing equations remain in tact because some of the governing equations, such as the continuity equation, are self-contained.
All of the tokens are then compiled into a single list, where each token in the tokenized equation is the index at which it occurs in this list.
For example, we have the following tokenization:
\vspace{-4mm}
\begin{equation*}
    \begin{split}
        \frac{\partial}{\partial t} u(x, t) &= Derivative(u(x,t), t) \\
        &= [Derivative, (, u, (, x, ,, t, ), ,, t, )] \\
        &= [6,0,3,15,0,16,33,14,1,33,1]
    \end{split}
\vspace{-4mm}
\end{equation*}

After each equation has been tokenized, the target time value is appended in tokenized form to the equation, and the total equation is padded with a placeholder token so that each text embedding is the same length.
Sampled values are truncated at 15 digits of precision.
Data handling code is adapted from PDEBench\cite{takamoto2022pdebench}.

\vspace{-4mm}
\subsection{Physics Informed Token Transformer}
\vspace{-4mm}
\label{sec:pitransformer}
The Physics Informed Token Transformer (PITT) utilizes tokenized equation information to construct an update operator $F_P$, similar to numerical integration techniques: $x_{t+1} = x_t + F_{P}(x_t)$.
We see in figure \ref{fig:pitt}, PITT takes in the numerical values and grid spacing, similar to operator learning architectures such as FNO, as well as the tokenized equation and the explicit time differential between simulation steps.
The tokenized equation is passed through a Multi Head Attention block seen in figure \ref{fig:token_transformer}.
In our case we use Self Attention\cite{cao_choose_2021}.
The tokens are shifted and scaled to be between -1 and 1 upon input, which significantly boosts performance.
This latent equation representation is then used to construct the keys and queries for a subsequent Multi Head Attention block that is used in conjunction with output from the underlying neural operator to construct the update values for the final input frame.
The time difference between steps is encoded, allowing use of arbitrary timesteps.
Intuitively, we can view the model as using a neural operator to passthrough the previous state, as well as calculate the update, like in numerical methods.
The tokenized information is then used to construct an analytically driven update operator that acts as a correction to the neural operator state update.
This intuitive understanding of PITT is explored with our 1D benchmarks. 

Two different embedding methods are used for the tokenized equations.
In the first method, the token attention block first embeds the tokens, $T$, as key, query, and values with learnable weight matrices: $T_1 = W_{T1} T$, $T_2 = W_{T2} T$, $T_3 = W_{T3} T$.
While this approach introduces unconventional correlations between tokens, only numerical values are modified in experiments, and so the correlation between numerical values is useful.
The second method uses standard fixed positional encoding\cite{vaswani_attention_2017} and lookup table embedding.
The standard approach does not introduce unconventional correlations between numerical values.
Dropout and Multi-head Self Attention ($SA$) are then used to compute a hidden representation: $T_h = Dropout(SA(T_1, T_2, T_3))$.
We use a single layer of self-attention for the tokens.
The update attention blocks seen in figure \ref{fig:linatt_update} then uses the token attention block output as queries and keys, the neural operator output as values, and embeds them using trainable matrices as $V_0 = W_X X_0$, $T_{h1} = W_{Th1} T_h$, $T_{h2} = W_{Th2} T_h$.
The output is passed through a fully connected projection layer to match the target output dimension.
This update scheme mimics numerical methods and is given in algorithm \ref{alg:pitt_numerical_update}.
\begin{algorithm}
\caption{PITT numerical update scheme}
\label{alg:pitt_numerical_update}
    \begin{algorithmic}
    \Require $V_0$, $T_{h1}$, $T_{h2}$, time $t$, $L$ layers
    \For{$l = 1,2,\ldots,L$}
        \State $X_l \gets Dropout(LA(T_{h1}, T_{h2}, V_{l-1})$
        \State $t_l \gets MLP\left(\frac{l\cdot t}{L}\right)$
        \State $V_l \gets V_{l-1} + MLP(\left[X_{l}, t_{l}\right])$
    \EndFor
    \end{algorithmic}
\end{algorithm}

A standard, fully connected multi-layer perceptron is used to calculate the update after concatenating the attention output with an embedding of the fractional timestep.
This block uses softmax-free Linear Attention ($LA$)\cite{cao_choose_2021}, computed as $\mathbf{z} = Q\left(\tilde{K}^T\tilde{V}\right)/n$.
$\tilde{K}$ and $\tilde{V}$ indicate instance normalization.
Note that the target time $t$ is incremented fractionally to more closely model numerical method updates.
Using multiple update layers and a fractional timestep is useful for long target times, such as steady-state or fixed-future type experiments.
Using a single update layer works will with small timesteps, such as predicting the next simulation step.

\vspace{-4mm}
\vspace{-4mm}
\section{Data Generation}
\vspace{-4mm}
In order to properly assess performance, multiple data sets that represent distinct challenges are used.
In the 1D case, we have the Heat equation, which is a linear parabolic equation, the KdV equation which is a nonlinear hyperbolic equation, and Burgers' equation.
In 2D, we have the Navier Stokes equations and the steady state Poisson equation.
Many parameters and forcing functions are sampled in order to generate large data sets.

\vspace{-4mm}
\subsection{Heat, Burgers', KdV Equations}
\vspace{-4mm}
Following the setup from \citet{brandstetter2022message}, we generate the 1D data.
In this case, a large number of sampled parameters allow us to generate many different initial conditions and forcing terms for each equation. In our case, $J=5$ and $L=16$.
\vspace{-4mm}
\begin{equation}
\left[\partial_t u + \partial_x\left(\alpha u^2 - \beta \partial_x u + \gamma \partial_{xx} u\right) \right]\left(t, x\right) = \delta\left(t, x\right)
\label{eq:1d_equation}
\vspace{-4mm}
\end{equation}
Where the forcing term is given by: $\delta\left(t, x\right) = \sum_{j=1}^JA_j \sin\left(\omega_j t + (2\pi l_j x)/L + \phi_j\right)$, and the initial condition is the forcing term at time $t=0$: $u\left(0, x\right) = \delta\left(0, x\right)$.
The parameters in the forcing term are sampled as follows: $A_j \sim \mathcal{U}(-0.5, 0.5), \quad \omega_j \sim \mathcal{U}(-0.4, 0.4), \quad l_j \sim \{1, 2, 3\}, \quad \phi_j \sim \mathcal{U}(0, 2\pi)$.
The parameters, $(\alpha, \beta, \gamma)$ of equation \ref{eq:1d_equation} can be set to define different, famous equations. 
When $\gamma = 0$, $\beta = 0$ we have the Heat equation, when only $\gamma = 0$ we have Burgers' equation, and when $\beta = 0$ we have the KdV equation.
Each equation has at least one parameter that we modify in order to generate large data sets.

%
%

For the Heat and Burgers' equations , we used diffusion values of $\beta \in \left\{0.01, 0.05, 0.1, 0.2, 0.5, 1 \right\}$.
For the Heat equation, we generated 10,000 simulations from each $\beta$ value for 60,000 total samples.
For Burgers' equation, we used advection values of $\alpha \in \left\{0.01, 0.05, 0.1, 0.2, 0.5, 1 \right\}$, and generated 2,500 simulations for each combination of values, for 90,000 total simulations.
For the KdV equation, we used an advection value of $\alpha=0.01$, with $\gamma \in \left\{2,4,6,8,10,12\right\}$, and generated 2,500 simulations for each parameter combination, for 15,000 total simulations.
The 1D equations text tokenization is padded to a length of 500.
Tokenized equations are long here due to the many sampled values.
\vspace{-4mm}
\subsection{Navier-Stokes Equation}
\vspace{-4mm}
In 2D, we use the incompressible, viscous Navier-Stokes equations in vorticity form, given in equation \ref{eq:ns}.
Data generation code was adapted from \citet{li2021fourier}.
\vspace{-4mm}
\begin{equation}
    \begin{split}
        \frac{\partial}{\partial t}w(x,t) + u(x,t)\cdot\nabla w(x,t) &= \nu\Delta w(x,t) + f(x) \\
        \nabla \cdot u(x,t) = 0, &\quad w(x,0) = w_0(x) \\
        f(x) &= A\left(\sin\left(2\pi(x_1 + x_2)\right) + \cos\left(2\pi(x_1+x_2)\right)\right)
    \end{split}
    \label{eq:ns}
                \vspace{-4mm}
\end{equation}
where $u(x,t)$ is the velocity field, $w(x,t) = \nabla \times u(x,t)$ is the vorticity, $w_0(x)$ is the initial vorticity, $f(x)$ is the forcing term, and $\nu$ is the viscosity parameter.
We use viscosities $\nu\in\left\{10^{-9}, 2\cdot 10^{-9}, 3\cdot 10^{-9}, \ldots, 10^{-8}, 2\cdot 10^{-8}, \ldots, 10^{-5} \right\}$ and forcing term amplitudes $A \in \left\{0.001, 0.002, 0.003, \ldots, 0.01\right\}$, for 370 total parameter combinations.
120 frames are saved over 30 seconds of simulation time.
The initial vorticity is sampled according to a gaussian random field.
For each combination of $\nu$ and $A$, 1 random initialization was used for the next-step and rollout experiments and 5 random initializations were used for the fixed-future experiments.
The tokenized equations are padded to a length of 100.
Simulations are run on a 1x1 unit cell with periodic boundary conditions.
The space is discretized with a 256x256 grid for numerical stability that is evenly downsampled to 64x64 during training and testing.

\vspace{-4mm}
\subsection{Steady-State Poisson Equation}
\vspace{-4mm}

\begin{figure}
    \centering
    \includegraphics[width=\linewidth]{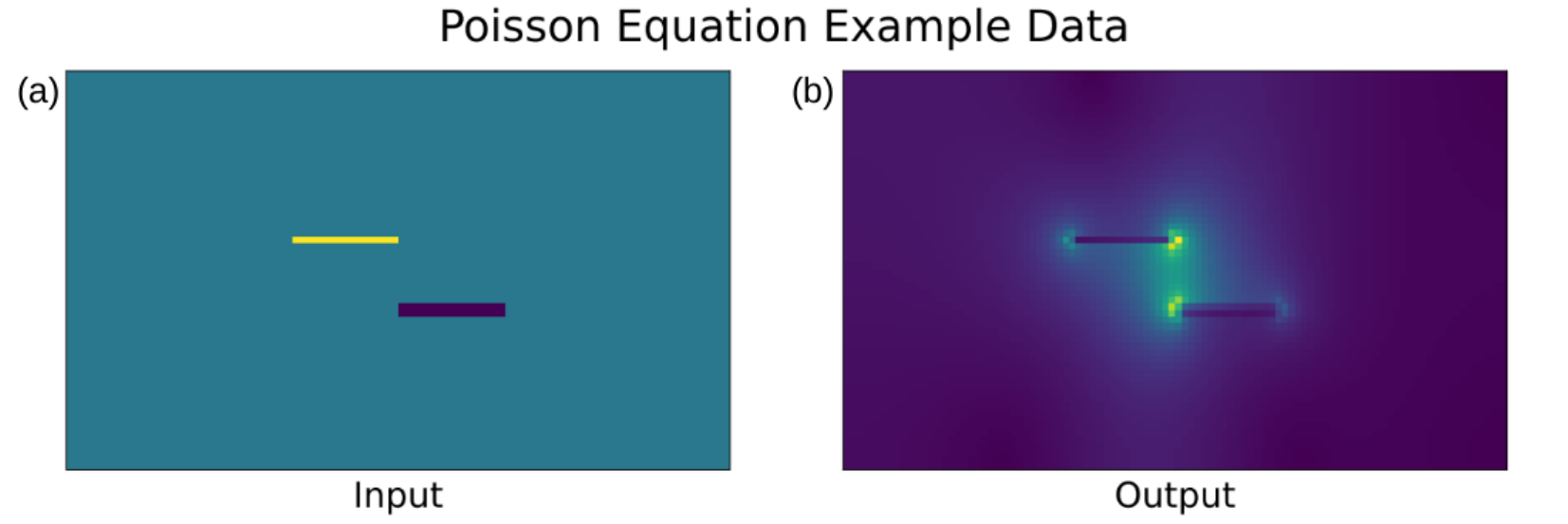}
    \caption{Example setup for the 2D Poisson equation. \textbf{a)} Input boundary conditions and geometry. \textbf{b)} Target electric field output.}
    \label{fig:poisson_equation}
\end{figure}

The last benchmark we perform is on the steady-state Poisson equation given in equation \ref{eq:poisson}.
\vspace{-4mm}
\vspace{-4mm}
\begin{equation}
    \nabla^2 u(x,y) = g(x,y)
    \label{eq:poisson}
\vspace{-4mm}
\end{equation}
where $u(x,y)$ is the electric potential, $-\nabla u(x,y)$ is the electric field, and $g(x,y)$ contains boundary condition and charge information.
The simulation cell is discretized with $100$ points in the horizontal direction and $60$ points in the vertical direction.
Capacitor plates are added with various widths, $x$ and $y$ positions, and charges.
An example of input and target electric field magnitude is given in figure \ref{fig:poisson_equation}.

This represents a substantially different task compared to previous benchmarks.
Due to the large difference between initial and final states, models must learn to extract significantly more information from provided input.
This benchmark also easily allows for testing how well models are able to learn Neumann, and generalize to different combinations of boundary conditions.
In two dimensions, we have four different boundaries on the simulation cell.
Each boundary takes either Dirichlet or Neumann boundary conditions, allowing for 16 different combinations.
In this case, since steady state is at infinite time, we pass the same time of 1 for each sample into the explicitly time-dependent models.
The tokenized equation and system parameters are padded to a length of 100.
Code is adapted from \citet{zaman_numerical_2022} for this case.

\vspace{-4mm}
\vspace{-4mm}
\section{Results}
\vspace{-4mm}
\label{sec:results}
We now compare PITT with both embedding methods against FNO, DeepONet, and OFormer on our various data sets.
$\dagger$ indicates our novel embedding method and * indicates standard embedding.
All experiments were run with five random splits of the data.
Reported results and shaded regions in plots are the mean and one standard deviation of each result, respectively.
Experiments were run with a 60-20-20 train-validation-test split.
Early stopping is also used, where the epoch with lowest validation loss is used for evaluation.
Note: parameter count represents total number of parameters.
In some cases PITT variants use a smaller underlying neural operator and have lower parameter count than the baseline model.
Hyperparameters for each experiment are given in the appendix.

\vspace{-4mm}
\subsection{1D Next-Step Prediction}
\vspace{-4mm}
\label{sec:1d_next_step_benchmarks}
Our 1D case is trained by using 10 frames of our simulation to predict the next frame.
The data is generated for four seconds, with 100 timesteps, and 100 grid points between 0 and 16.
The final time is $T=4s$.
Specifically, the task is to learn the operator $\mathcal{G}_{\theta}:a(\cdot,t_i)|_{i\in[n-9, n]} \to u(\cdot,t_j)|_{j=n+1}$ where $n \in [10, 100]$.
A total of 1,000 sampled equations were used in the training set, with 90 frames for each equation.
Data was split such that samples from the same equation and forcing term did not appear in the training and test sets.
We see PITT significantly outperforms all of the baseline models across all equations for both embedding methods.
Although the lower error often resulted in unstable autoregressive rollout, PITT variants have also outperformed their baseline counterparts when simply trained to minimum error.
Additionally, PITT is able to improve performance with fewer parameters than FNO, and a comparable number of parameters to both OFormer and DeepONet.
Notably, PITT uses a single attention head and single multi-head attention block for the multi-head and linear attention blocks in this experiment.
\begin{table}
    \centering
    \resizebox{\linewidth}{!}{
    \begin{tabular}{c|c | ccc}
         Model & Parameter Count & Heat & Burgers' & KdV\\
         \hline
         FNO & 2.4M & 4.80 $\pm$ 0.18 & 8.22 $\pm$ 0.37 & 11.28 $\pm$ 0.43\\
         PITT FNO\textsuperscript{$\dagger$} (Ours) & 0.2M & \textbf{0.38 $\pm$ 0.02} & \textbf{0.23 $\pm$ 0.06} & \textbf{8.77 $\pm$ 0.20}\\
         PITT FNO\textsuperscript{*} (Ours) & 0.4M & \textbf{0.38 $\pm$ 0.01} & \textbf{0.66 $\pm$ 0.07} & \textbf{8.68 $\pm$ 0.21}\\
         \hdashline
         OFormer & 3.0M & 1.44 $\pm$ 0.17 & 4.32 $\pm$ 0.35 & 4.36 $\pm$ 0.21\\
         PITT OFormer\textsuperscript{$\dagger$} (Ours) & 0.4M & \textbf{0.06 $\pm$ 0.03} & \textbf{0.22 $\pm$ 0.02} & \textbf{0.46 $\pm$ 0.03}\\
         PITT OFormer\textsuperscript{*} (Ours) & 0.5M & \textbf{0.23 $\pm$ 0.13} & \textbf{0.24 $\pm$ 0.04} & \textbf{0.47 $\pm$ 0.02}\\
         \hdashline
         DeepONet & 0.2M & 0.68 $\pm$ 0.06 & 2.14 $\pm$ 0.17 & 9.22 $\pm$ 0.31\\
         PITT DeepONet\textsuperscript{$\dagger$} (Ours) & 0.2M & \textbf{0.03 $\pm$ 0.01} & \textbf{0.24 $\pm$ 0.05} & \textbf{1.78 $\pm$ 0.10}\\
         PITT DeepONet\textsuperscript{*} (Ours) & 0.3M & \textbf{0.02 $\pm$ 0.01} & \textbf{0.21 $\pm$ 0.06} & \textbf{8.25 $\pm$ 0.28}\\
    \end{tabular}}
    \caption[H!]{Mean Absolute Error (MAE) $\times 10^{-3}$ for 1D benchmarks. \textbf{Bold} indicates best performance.}
    \label{tab:1d_results}
\end{table}

The effect of the neural operator and token transformer modules in PITT can be easily decomposed and analyzed by returning the passthrough and update separately, instead of their sum (Figure \ref{fig:PITT_framework}).
Using the pretrained PITT FNO from above, a sample is predicted for the 1D Heat equation.
We see the decomposition in figure \ref{fig:pitt_decomp_novel} and figure \ref{fig:pitt_decomp_standard} in the appendix.
\begin{figure}[ht]
    \centering
    \includegraphics[width=0.95\linewidth]{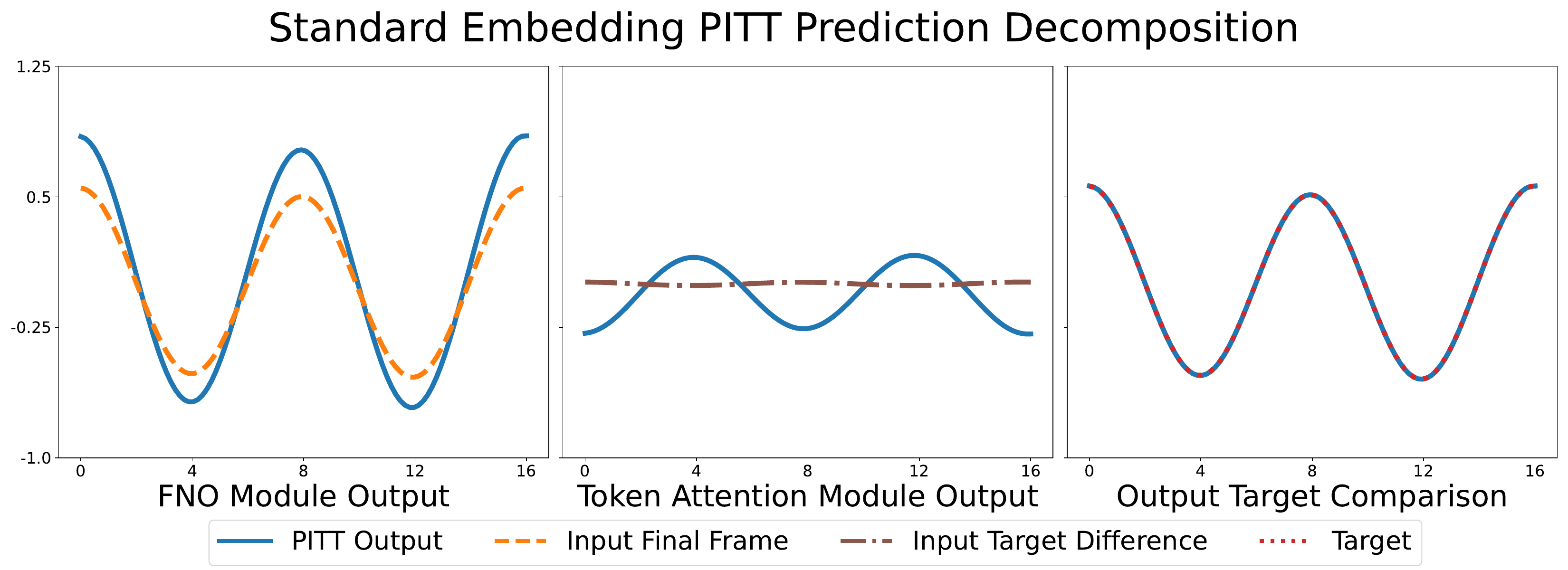}
    \caption{PITT FNO prediction decomposition for 1D Heat equation. \textbf{Left:} The FNO module of PITT predicts a large change to the final frame of input data. \textbf{Middle} The numerical update block corrects the FNO output. \textbf{Right} The combination of FNO and numerical update block output very accurately predicts the next step.}
    \label{fig:pitt_decomp_novel}
\end{figure}

Interestingly, the underlying FNO has learned to overestimate the passthrough of the data in both cases.
The token attention and numerical update modules have learned a correction to the FNO output, as expected.

\vspace{-4mm}
\subsection{1D Fixed-Future Prediction}
\vspace{-4mm}
\label{sec:1d_fixed_future_benchmarks}
In this 1D benchmark, each model is trained on all three equations simultaneously, and performance is compared against training on single equations.
Results are shown in table \ref{tab:multi_1d_results}.
The first 10 frames of each equation are used as input to predict the last frame of each simulation.
In total, 5,000 samples from each equation were used for both single equation and multiple equation training.
Models trained on the combined data sets are then tested on data from each equation individually.
For PITT FNO and PITT OFormer, we see that training on the combined equations using our novel embedding method has best performance across all data sets.
Additionally, for PITT FNO and PITT DeepONet, training using our standard embedding method acheivs best performance across all data sets.
This shows PITT is able to improve neural operator generalization across different systems.
Interestingly, we see also improvement in FNO and OFormer when training using the combined data sets.

\begin{table}
    \centering
    \resizebox{\linewidth}{!}{
    \begin{tabular}{c|c | ccc}
         Model & Parameter Count & Heat & Burgers' & KdV \\ 
         \hline
         FNO & 2.4M & 0.439 $\pm$ 0.005 & 0.528 $\pm$ 0.019 & 0.404 $\pm$ 0.004 \\
         FNO Multi & 2.4M & 0.239 $\pm$ 0.002 & 0.285 $\pm$ 0.001 & 0.329 $\pm$ 0.002 \\
         \hdashline
         PITT FNO\textsuperscript{$\dagger$} (Ours) & 0.2M & 0.177 $\pm$ 0.002 & 0.211 $\pm$ 0.005 & 0.220 $\pm$ 0.007 \\
         PITT FNO\textsuperscript{$\dagger$} Multi (Ours) & 0.2M & \textbf{0.120 $\pm$ 0.002} & \textbf{0.133 $\pm$ 0.002} & \textbf{0.165 $\pm$ 0.005} \\
         \hdashline
         PITT FNO\textsuperscript{*} (Ours) & 0.3M & 0.158 $\pm$ 0.003 & 0.205 $\pm$ 0.003 & 0.194 $\pm$ 0.005 \\
         PITT FNO\textsuperscript{*} Multi (Ours) & 0.3M & \textbf{0.124 $\pm$ 0.005} & \textbf{0.135 $\pm$ 0.019} & \textbf{0.166 $\pm$ 0.004} \\
         \hline
         OFormer & 3.0M & 0.154 $\pm$ 0.003 & 0.192 $\pm$ 0.004 & 0.244 $\pm$ 0.004 \\
         OFormer Multi & 3.0M & 0.150 $\pm$ 0.003 & 0.166 $\pm$ 0.002 & 0.210 $\pm$ 0.002 \\
         \hdashline
         PITT OFormer\textsuperscript{$\dagger$} (Ours) & 0.2M & 0.202 $\pm$ 0.008 & 0.222 $\pm$ 0.007 & 0.233 $\pm$ 0.004 \\
         PITT OFormer\textsuperscript{$\dagger$} Multi (Ours) & 0.2M & \textbf{0.142 $\pm$ 0.004} & \textbf{0.160 $\pm$ 0.005} & \textbf{0.191 $\pm$ 0.006} \\
         \hdashline
         PITT OFormer\textsuperscript{*}(Ours) & 0.3M & 0.201 $\pm$ 0.006 & 0.228 $\pm$ 0.008 & 0.232 $\pm$ 0.006 \\
         PITT OFormer\textsuperscript{*} Multi (Ours) & 0.3M & 0.154 $\pm$ 0.003 & 0.170 $\pm$ 0.004 & \textbf{0.200 $\pm$ 0.004} \\
         \hline
         DeepONet & 0.2M & 0.240 $\pm$ 0.003 & 0.420 $\pm$ 0.008 & 0.519 $\pm$ 0.008 \\
         DeepONet Multi & 0.2M & 0.608 $\pm$ 0.009 & 0.609 $\pm$ 0.006 & 0.749 $\pm$ 0.014 \\
         \hdashline
         PITT DeepONet\textsuperscript{$\dagger$} (Ours) & 0.3M & \textbf{0.185 $\pm$ 0.002} & 0.355 $\pm$ 0.005 & 0.488 $\pm$ 0.007 \\
         PITT DeepONet\textsuperscript{$\dagger$} Multi (Ours) & 0.3M & 0.214 $\pm$ 0.009 & \textbf{0.330 $\pm$ 0.006} & 0.488 $\pm$ 0.007 \\
         \hdashline
         PITT DeepONet\textsuperscript{*} (Ours) & 0.4M & 0.195 $\pm$ 0.006 & 0.320 $\pm$ 0.017 & \textbf{0.482 $\pm$ 0.009} \\
         PITT DeepONet\textsuperscript{*} Multi (Ours) & 0.4M & \textbf{0.187 $\pm$ 0.003} & \textbf{0.270 $\pm$ 0.008} & \textbf{0.481 $\pm$ 0.008} \\
    \end{tabular}}
    \caption[H!]{Mean Absolute Error (MAE) $\times 10^{-3}$ for 1D benchmarks. \textbf{Bold} indicates best performance.}
    \label{tab:multi_1d_results}
\end{table}

\vspace{-4mm}
\subsection{2D Benchmarks}
\vspace{-4mm}
The 2D benchmarks provided here provide a wider array of settings and tests for each model.
In the next-step training and rollout test experiment, we used 200 equations, a single random initialization for each equation, and the entire 121 step trajectory for the data set.
The final time is $T=30s$.
Similar to the 1D case, we are learning the operator $\mathcal{G}_{\theta}:a(\cdot,t_i)|_{i=n} \to u(\cdot,t_j)|_{j=n+1}$ where $n \in [0, 119]$.
This benchmark is especially challenging for two reasons.
First, there are viscosity and forcing term amplitude combinations in the test set that the model has not trained on.
Second, rollout is done starting from only the initial condition, and models are trained to predict the next step using a single snapshot.
This limits the time evolution information available to models during training.
Although the baseline models perform comparably to PITT variants in terms of error, we note that PITT shows improved accuracy for all variants, and in many cases lower error led to unstable rollout, like in the 1D cases.
Despite this, PITT has much better rollout error accumulation, seen in table \ref{tab:rollout_results}.
Further analysis of PITT FNO attention maps from this experiment is given in the appendix in figures \ref{fig:old_pitt_token_exploration}, \ref{fig:old_pitt_token_exploration_time}, \ref{fig:new_pitt_token_exploration}, and \ref{fig:new_pitt_token_exploration_time}.
The attention maps show PITT FNO is able to extract physically relevant information from the governing equations.

For the steady-state Poisson equation, for a given set of boundary conditions we learn the operator, $\mathcal{G}_{\theta}:a \to u$, with Boundary conditions: $ u(x)=g(x), \forall x \in \partial \Omega_0 $ and $ \mathbf{\hat{n}}\nabla u(x)= f(x), \forall x \in \partial \Omega_1$.
The primary challenge here is in learning the effect of boundary conditions.
Dirichlet boundary conditions are constant, only requiring passing through initial values at the boundary for accurate prediction, but Neumann boundary conditions lead to boundary values that must be learned from the system.
Standard neural operators do not offer a way to easily encode this information without modifying the initial conditions, while PITT uses a text encoding of each boundary condition, as outlined in equation tokenization.
PITT is able to learn boundary conditions through the text embedding, and performs approximately an order of magnitude better, with the standard embedding improving over our novel embedding by an average of over 50\%.
5,000 samples were used during training with random data splitting.
All combinations of boundary conditions appear in both the train and test sets.
Prediction error plots for our models on this data set are given in the appendix in figures \ref{fig:old_poisson_error_comp} and \ref{fig:new_poisson_error_comp}.

\begin{table}
    \centering
    \begin{tabular}{c|c | cc}
         Model & Parameter Count & Navier-Stokes & Poisson \\
         \hline
         FNO & 2.1M/8.5M & 5.24 $\pm$ 0.30 &  9.79 $\pm$ 0.12\\
         PITT FNO\textsuperscript{$\dagger$} (Ours) & 1.0M/4.2M & \textbf{5.07 $\pm$ 0.30} & \textbf{1.15 $\pm$ 0.17} \\
         PITT FNO\textsuperscript{*} (Ours) & 1.7M/4.0M & \textbf{5.18 $\pm$ 0.29} & \textbf{0.85 $\pm$ 0.05} \\
         \hdashline
         OFormer & 1.0M/0.2M & \textbf{10.07 $\pm$ 0.94} & 9.98 $\pm$ 0.11 \\
         PITT OFormer\textsuperscript{$\dagger$} (Ours) & 0.9M/2.0M & 14.63 $\pm$ 3.42 & \textbf{0.69 $\pm$ 0.38} \\
         PITT OFormer\textsuperscript{*} (Ours) & 1.2M/2.2M & 20.54 $\pm$ 0.94 & \textbf{0.33 $\pm$ 0.02} \\
         \hdashline
         DeepONet & 0.3M/0.4M & 7.06 $\pm$ 0.32 & 25.20 $\pm$ 0.22 \\
         PITT DeepONet\textsuperscript{$\dagger$} (Ours) & 1.7M/3.2M & \textbf{7.01 $\pm$ 0.31} & \textbf{1.50 $\pm$ 1.40}\\
         PITT DeepONet\textsuperscript{*} (Ours) & 1.5M/2.5M & \textbf{7.01 $\pm$ 0.32} & \textbf{0.53 $\pm$ 0.04}\\
    \end{tabular}
    \label{tab:2d_eq_results}
    \caption[top]{Mean Absolute Error (MAE) $\times 10^{-3}$ for 2D benchmarks. \textbf{Bold} indicates best performance. Although PITT variants have overlapping error bars with the base model in the Navier-Stokes benchmark, the PITT variant had lower error on all but one random split of the data for PITT FNO, and every random split for PITT DeepONet.}
\end{table}

Lastly, similar to experiments in both \citet{li2021fourier} and \citet{li2022transformer}, we can use our models to use the first $10$ seconds of data to predict a fixed, future timestep.
Including the initial condition, we use 41 frames to predict a single, future frame.
In this case, we predict the system state at $20$ and $30$ seconds in two separate experiments.
For this experiment, we are learning the operator $\mathcal{G}_{\theta}:u(\cdot,t)|_{t\in[0, 10]} \to u(\cdot,t)|_{t=20,30}$.
We shuffle the data such that forcing term amplitude and viscosity combinations appear in both the training and test set, but initial conditions do not appear in both.
Our setup is more difficult than in previous works because we are using multiple forcing term amplitudes and viscosities.
The results are given in table \ref{tab:2d_fixed_future_results}, where we see PITT variants outperform the baseline model for both embedding methods.
Example predictions are given in the appendix in figures \ref{fig:old_ff} and \ref{fig:new_ff}.

\begin{table}
    \centering
    \begin{tabular}{c|c|cc}
         Model &  Parameter Count & T=20 & T=30 \\
         \hline
         FNO & 0.3M & 4.44 $\pm$ 0.05 & 8.11 $\pm$ 0.08 \\
         PITT FNO\textsuperscript{$\dagger$} (Ours) & 0.3M & \textbf{4.06 $\pm$ 0.13} & \textbf{7.26 $\pm$ 0.16} \\
         PITT FNO\textsuperscript{*} (Ours) & 1.6M & \textbf{4.02 $\pm$ 0.03} & \textbf{7.46 $\pm$ 0.09} \\
         \hdashline
         OFormer & 0.3M & 5.91 $\pm$ 0.16 & 8.83 $\pm$ 0.15 \\
         PITT OFormer\textsuperscript{$\dagger$} (Ours) & 0.5M & \textbf{5.64 $\pm$ 0.16} & \textbf{8.38 $\pm$ 0.07} \\
         PITT OFormer\textsuperscript{*} (Ours) & 1.6M & \textbf{5.75$\pm$ 0.20} & \textbf{8.54 $\pm$ 0.09} \\
         \hdashline
         DeepONet & 0.3M & 10.28 $\pm$ 0.11 & 14.69 $\pm$ 0.19 \\
         PITT DeepONet\textsuperscript{$\dagger$} (Ours) & 0.5M & \textbf{8.96 $\pm$ 0.10} & \textbf{12.35 $\pm$ 0.09}\\
         PITT DeepONet\textsuperscript{*} (Ours) & 1.1M & \textbf{8.52 $\pm$ 0.08} & \textbf{11.33 $\pm$ 0.12} \\
    \end{tabular}
    \caption{Mean Absolute Error (MAE) $\times 10^{-2}$ for 2D Fixed-Future Benchmarks. \textbf{Bold} indicates best performance.}
    \label{tab:2d_fixed_future_results}
\end{table}



\vspace{-4mm}
\subsection{Rollout}
\begin{figure}[H]
    \centering
    \includegraphics[width=0.95\linewidth, height=4.3cm]{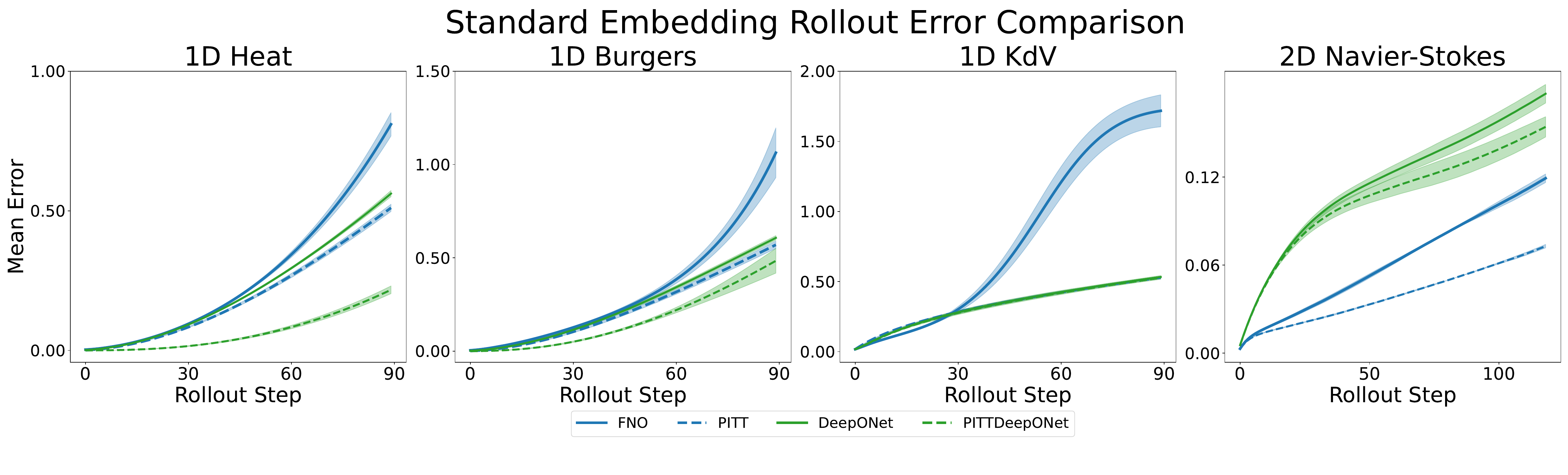}
    \caption{Error accumulation for rollout experiments using standard embedding.}
    \label{fig:1d_rollout_standard}
\end{figure}

\vspace{-4mm}
\label{sec:rollout_results}
An important test of viability for operator learning models as surrogate models is how error accumulates over time.
In real-world predictions, we often must autoregressively predict into the future, where training data is not available.
OFormer results are not presented here due to instability in autoregressive rollout.
We see in table \ref{tab:rollout_results} that our PITT variants shows significantly less final error at large rollout times for all time-dependent data sets, with the exception of PITT DeepONet using our novel embedding method when compared to the baseline model on KdV.
Error accumulation is shown in figure \ref{fig:1d_rollout_standard} for standard embedding, where PITT shows both lower final error and improved total error accumulation.
The novel embedding error accumulation plot is given in the appendix in figure \ref{fig:1d_rollout_novel}.
In these experiments, we used the models trained in the next-step fashion from section our 1D benchmarks.
We start with the first 10 frames from each trajectory in the test set for the 1D data sets and the only initial condition for the 2D test data set and autoregressively predict the entire rollout.

\begin{table}[h]
    \centering
    \resizebox{\linewidth}{!}{
    \begin{tabular}{c|c|cccc}
         Model &  Parameter Count & 1D Heat & 1D Burgers' & 1D KdV & 2D NS \\
         \hline
         FNO & 2.4M/2.1M & 0.810 $\pm$ 0.042 & 1.063 $\pm$ 0.133 & 1.718 $\pm$ 0.114 & 0.125 $\pm$ 0.006 \\
         PITT FNO\textsuperscript{$\dagger$} (Ours) & 0.2M/1.0M & \textbf{0.483 $\pm$ 0.015} & \textbf{0.351 $\pm$ 0.19} & \textbf{0.555 $\pm$ 0.050} & \textbf{0.065 $\pm$ 0.003} \\
         PITT FNO\textsuperscript{*} (Ours) & 0.2M/1.0M & \textbf{0.511 $\pm$ 0.013} & \textbf{0.570 $\pm$ 0.020} & \textbf{0.529 $\pm$ 0.008} & \textbf{0.073 $\pm$ 0.004} \\
         \hdashline
         OFormer & N/A & N/A & N/A & N/A & N/A \\
         PITT OFormer\textsuperscript{$\dagger$} (Ours) & N/A & N/A & N/A & N/A & N/A \\
         PITT OFormer\textsuperscript{*} (Ours) & N/A & N/A & N/A & N/A & N/A \\
         \hdashline
         DeepONet & 0.2M/0.8M & 0.562 $\pm$ 0.011 & 0.607 $\pm$ 0.0121 & 0.533 $\pm$ 0.010 & 0.179 $\pm$ 0.005 \\
         PITT DeepONet\textsuperscript{$\dagger$} (Ours) & 0.4M/1.7M & \textbf{0.404 $\pm$ 0.100} & \textbf{0.536 $\pm$ 0.105} & 0.699 $\pm$ 0.048 & \textbf{0.154 $\pm$ 0.008}\\
         PITT DeepONet\textsuperscript{*} (Ours) & 0.3M/1.7M & \textbf{0.217 $\pm$ 0.0138} & \textbf{0.484 $\pm$ 0.066} & \textbf{0.526 $\pm$ 0.011} & \textbf{0.157 $\pm$ 0.012}\\
    \end{tabular}}
    \caption{Final Mean Absolute Error (MAE) for rollout experiments. \textbf{Bold} indicates best performance when comparing base models to their PITT version. OFormer is omitted due to instability during rollout. Standard embedding PITT DeepONet is bolded here because it outperforms DeepONet for every random split of the data.}
    \label{tab:rollout_results}
\end{table}


\begin{figure}
    \centering
    \includegraphics[width=0.9\linewidth]{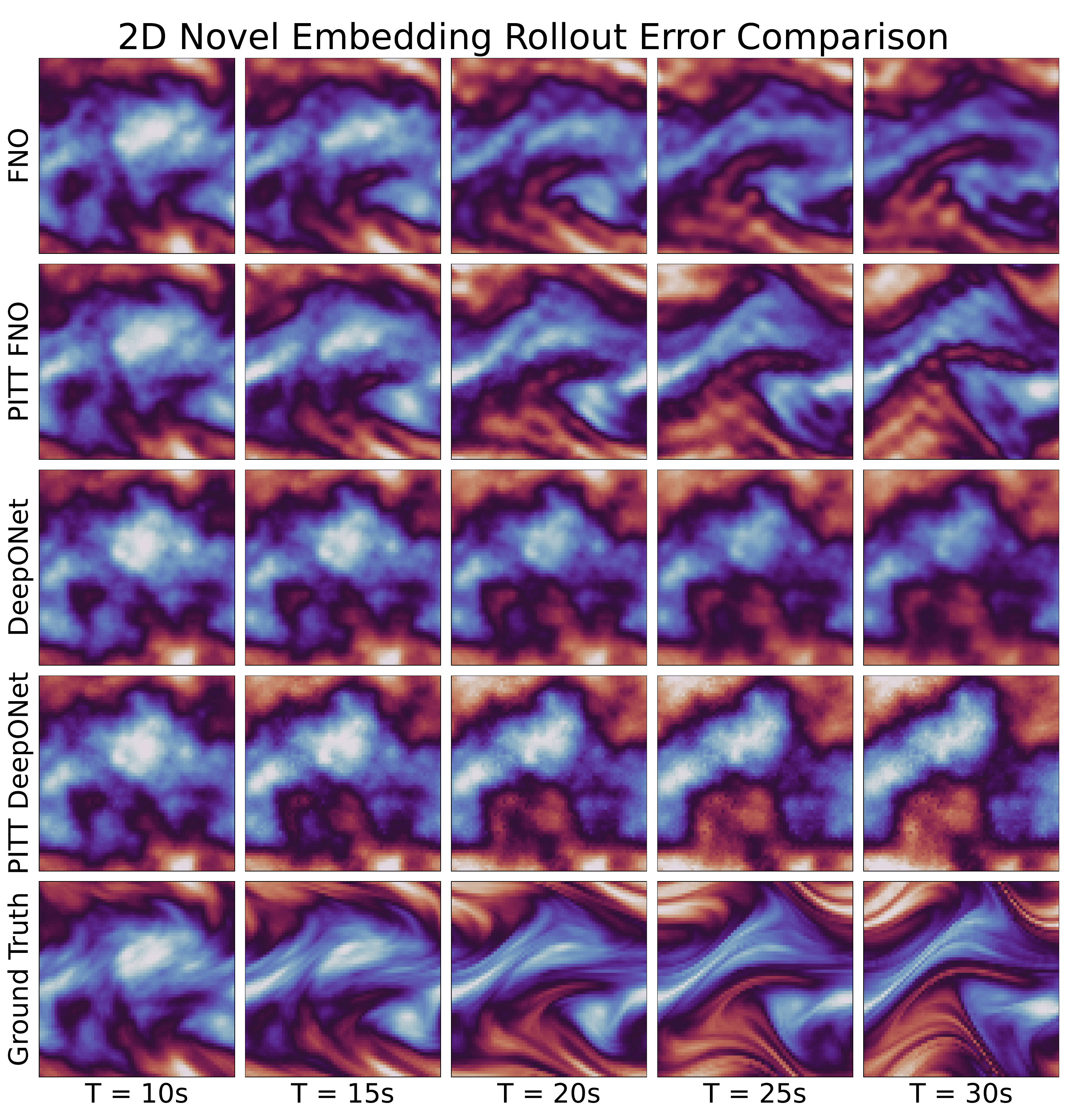}
    \caption{Rollout results for 2D Navier Stokes using our novel embedding method.}
    \label{fig:2d_rollout_ns}
\end{figure}

A visualization of 2D rollout for our novel embedding method is given in figure \ref{fig:2d_rollout_ns}.
At long rollout times, especially $T=25s$ and $T=30s$, PITT FNO is able to accurately predict large-scale features, with accurate prediction of some of the smaller scale features.
FNO, on the other hand, has begun to predict noticeably different features from the ground truth, and does not match small-scale features well.
Similarly, PITT DeepONet is able to approximately match large-scale features in magnitude (lighter color), whereas DeepONet noticably differs even at large scales.
1D rollout comparison plots are given in the appendix.

\vspace{-4mm}
\vspace{-4mm}
\section{Conclusion}
\vspace{-4mm}
This work introduces a novel transformer-based architecture, PITT, that learns analytically-driven numerical update operators from governing equations.
A novel equation embedding method is developed and compared against standard positional encoding and embedding.
PITT is able to learn physically relevant information from tokenized equations and outperforms baseline neural operators on a wide variety of challenging 1D and 2D benchmarks.
We have also found our baseline models and their PITT variants with both embedding strategies have lower time-to-solution than the numerical methods used for data generation.
Details of the timing experiment and results are given in the appendix in tables \ref{tab:ns_pred_time} and \ref{tab:ff_pred_time} for our 1D next-step and fixed future experiments, respectively.
Future work includes benchmarking on 3D systems, more effective tokenization and efficient embedding, as our novel method uses a naive approach that introduces unconventional correlation between tokens, but standard positional encoding and embedding does not use useful correlation between tokens.
Additionally, the current experiments have redundancy in equations as only system parameters such as viscosity vary.
Testing on multiple systems simultaneously would serve as a test for PITT's generalization capability.
In addition, other works\cite{li2021fourier} have used recurrent rollout prediction as well as training rollout trajectories, which we have currently have not evaluated.
These strategies can be employed to help stabilize rollout predictions.
\section{Supplementary Material}
Supplementary material contains training hyperparameters, analysis of PITT attention maps, and further exploration of results.
\section{Acknowledgements}
This material is based upon work supported by the National Science Foundation under Grant No. 1953222.

\section{Author Declaration}
The authors have no conflicts to disclose.

\section{Author Contributions}
Cooper Lorsung: Conceptualization, Methodology, Software, Validation, Formal Analysis, Investigation, Data Curation, Writing - Original Draft.
Zijie Li: Conceptualization, Writing - Original Draft, Writing - Review \& Editing.
Amir Barati Farimani: Conceptualization, Resources, Writing - Review \& Editing, Supervision, Funding acquisition.

\section{Data Availability}
Data and code are available at \href{https://github.com/BaratiLab/PhysicsInformedTokenTransformer}{https://github.com/BaratiLab/PhysicsInformedTokenTransformer}

\bibliography{the}

\appendix

\noindent
\Huge
\textbf{Appendix}
\normalsize
\section{Experimental Details}
\label{app:exp_details}

Training and model hyperparameters are given here.
In all cases, 3 encoding and decoding convolutional layers were used for FNO.
FNO and DeepONet used a step schedule for learning rate during training.
PITT variants and OFormer used a One Cycle Learning rate during training.
All models used the Adam optimizer and L1 loss function for all experiments.

\subsection{1D Next-Step Training Details}
\label{app:1d_params}

All models were trained for 200 epochs on all of the 1D data sets.
\begin{table}[H]
    \centering
    \resizebox{\linewidth}{!}{
    \begin{tabular}{c|c|cccccc}
         Model & Data Set & Batch Size & Learning Rate & Weight Decay & Dropout & Scheduler Step & Scheduler $\gamma$ \\
         \hline
         FNO & Heat & 32 & 1E-3 & 1E-8 & 0.1 & 50 & 0.5 \\
         FNO & Burger's & 32 & 1E-3 & 1E-8 & 0.1 & 50 & 0.5 \\
         FNO & KdV & 4 & 1E-3 & 1E-8 & 0.1 & 50 & 0.5 \\
         PITT FNO\textsuperscript{$\dagger$} & Heat & 128 & 1E-3 & 1E-5 & 0.0 & N/A & N/A \\
         PITT FNO\textsuperscript{$\dagger$} & Burger's & 128 & 1E-3 & 1E-5 & 0.0 & N/A & N/A \\
         PITT FNO\textsuperscript{$\dagger$} & KdV & 256 & 1E-3 & 1E-2 & 0.0 & N/A & N/A \\
         PITT FNO\textsuperscript{*} & Heat & 32 & 1E-3 & 1E-5 & 0.0 & N/A & N/A \\
         PITT FNO\textsuperscript{*} & Burger's & 16 & 1E-4 & 1E-4 & 0.0 & N/A & N/A \\
         PITT FNO\textsuperscript{*} & KdV & 128 & 1E-3 & 1E-2 & 0.0 & N/A & N/A \\
         OFormer & Heat & 32 & 1E-3 & 1E-6 & 0.0 & N/A & N/A \\
         OFormer & Burgers & 32 & 1E-3 & 1E-6 & 0.0 & N/A & N/A \\
         OFormer & KdV & 128 & 1E-3 & 1E-2 & 0.0 & N/A & N/A \\
         PITT OFormer\textsuperscript{$\dagger$} & Heat & 32 & 1E-3 & 1E-6 & 0.0 & N/A & N/A \\
         PITT OFormer\textsuperscript{$\dagger$} & Burgers & 32 & 1E-3 & 1E-6 & 0.0 & N/A & N/A \\
         PITT OFormer\textsuperscript{$\dagger$} & KdV & 32 & 1E-3 & 1E-6 & 0.0 & N/A & N/A \\
         PITT OFormer\textsuperscript{*} & Heat & 32 & 1E-3 & 1E-5 & 0.0 & N/A & N/A \\
         PITT OFormer\textsuperscript{*} & Burgers & 32 & 1E-3 & 1E-6 & 0.0 & N/A & N/A \\
         PITT OFormer\textsuperscript{*} & KdV & 32 & 1E-3 & 1E-6 & 0.0 & N/A & N/A \\
         DeepONet & Heat & 128 & 1E-3 & 1E-2 & N/A & 20 & 0.5 \\
         DeepONet & Burgers & 64 & 1E-3 & 1E-1 & N/A & 20 & 0.5 \\
         DeepONet & KdV & 32 & 1E-3 & 1E-1 & N/A & 20 & 0.5\\
         PITT DeepONet\textsuperscript{$\dagger$} & Heat & 32 & 1E-4 & 1E-4 & 0.2 & N/A & N/A \\
         PITT DeepONet\textsuperscript{$\dagger$} & Burgers & 128 & 1E-4 & 1E-4 & 0.2 & N/A & N/A \\
         PITT DeepONet\textsuperscript{$\dagger$} & KdV & 64 & 1E-3 & 1E-8 & 0.0 & N/A & N/A \\
         PITT DeepONet\textsuperscript{*} & Heat & 16 & 1E-4 & 1E-4 & 0.1 & N/A & N/A \\
         PITT DeepONet\textsuperscript{*} & Burgers & 128 & 1E-4 & 1E-4 & 0.2 & N/A & N/A \\
         PITT DeepONet\textsuperscript{*} & KdV & 64 & 1E-3 & 1E-8 & 0.0 & N/A & N/A \\
    \end{tabular}}
    \caption{Training Hyperparameters for 1D Next-Step Experiments}
    \label{tab:pitt_1d_training_hyperparams}
\end{table}

\begin{table}[H]
    \centering
    \resizebox{\linewidth}{!}{
    \begin{tabular}{c|c|cccccc}
         Model & Data Set & Hidden Dimension & Numerical Layers & Heads & FNO Modes \\
         \hline
         FNO & Heat & 256 & N/A & N/A & 8 \\
         FNO & Burger's & 256 & N/A & N/A & 8 \\
         FNO & KdV & 256 & N/A & N/A & 8 \\
         PITT FNO\textsuperscript{$\dagger$} & Heat & 64 & 1 & 1 & 4 \\
         PITT FNO\textsuperscript{$\dagger$} & Burger's & 64 & 1 & 1 & 4 \\
         PITT FNO\textsuperscript{$\dagger$} & KdV & 64 & 1 & 1 & 4 \\
         PITT FNO\textsuperscript{*} & Heat & 64 & 1 & 1 & 4 \\
         PITT FNO\textsuperscript{*} & Burger's & 64 & 1 & 1 & 4 \\
         PITT FNO\textsuperscript{*} & KdV & 64 & 1 & 1 & 4 \\
    \end{tabular}}
    \caption{FNO Hyperparameters for 1D Next-Step Experiments}
    \label{tab:pitt_1d_model_hyperparams}
\end{table}
\begin{table}[H]
    \centering
    \resizebox{\linewidth}{!}{
    \begin{tabular}{c|c|ccccccccccc}
         Model & Data Set & Hidden Dim. & Numerical Layers & Heads & Input Embedding Dim. & Output Embedding Dim. & Encoder Depth & Decoder Depth & Latent Channels & Encoder Resolution & Decoder Resolution  & Scale\\
         \hline
         OFormer & Heat & N/A & N/A & 1 & 64 & 256 & 2 & 2 &256 & 1024 & 1024 & 1 \\
         OFormer & Burgers & N/A & N/A & 1 & 64 & 256 & 2 & 2 &256 & 1024 & 1024 & 1 \\
         OFormer & KdV & N/A & N/A & 1 & 64 & 256 & 2 & 2 &256 & 1024 & 1024 & 1 \\
         PITT OFormer\textsuperscript{$\dagger$} & Heat & 64 & 1 & 1 & 64 & 256 & 2 & 2 & 256 & 128 & 256 & 1\\
         PITT OFormer\textsuperscript{$\dagger$} & Burgers & 64 & 1 & 1 & 64 & 256 & 2 & 2 & 256 & 128 & 256 & 1\\
         PITT OFormer\textsuperscript{$\dagger$} & KdV & 64 & 1 & 1 & 64 & 256 & 2 & 2 & 256 & 128 & 256 & 1\\
         PITT OFormer\textsuperscript{*} & Heat & 64 & 1 & 1 & 64 & 256 & 2 & 2 & 256 & 128 & 256 & 1 \\
         PITT OFormer\textsuperscript{*} & Burgers & 64 & 1 & 1 & 64 & 256 & 2 & 2 & 256 & 128 & 256 & 1 \\
         PITT OFormer\textsuperscript{*} & KdV & 64 & 1 & 1 & 64 & 256 & 2 & 2 & 256 & 128 & 256 & 1\\
    \end{tabular}}
    \caption{OFormer Hyperparameters for 1D Next-Step Experiments}
    \label{tab:pitt_1d_model_hyperparams}
\end{table}
\begin{table}[H]
    \centering
    \resizebox{\linewidth}{!}{
    \begin{tabular}{c|c|cccccccc}
         Model & Data Set & Hidden Dimension & Numerical Layers & Heads & Branch Net & Trunk Net & Activation & Initializer \\
         \hline
         DeepONet & Heat & N/A & N/A & N/A & [10,256,256] & [100,256,256] & silu & Glorot Normal \\
         DeepONet & Burgers & N/A & N/A & N/A & [10,256,256] & [100,256,256] & silu & Glorot Normal \\
         DeepONet & KdV & N/A & N/A & N/A & [10,256,256] & [100,256,256] & silu & Glorot Normal \\
         PITT DeepONet\textsuperscript{$\dagger$} & Heat & 128 & 1 & 1 & [10,128,128] & [100,128,128] & silu & Glorot Normal \\
         PITT DeepONet\textsuperscript{$\dagger$} & Burgers & 128 & 1 & 1 & [10,128,128] & [100,128,128] & silu & Glorot Normal \\
         PITT DeepONet\textsuperscript{$\dagger$} & KdV & 128 & 1 & 1 & [10,128,128] & [100,128,128] & silu & Glorot Normal \\
         PITT DeepONet\textsuperscript{*} & Heat & 64 & 1 & 1 & [10,128,128] & [100,128,128] & silu & Glorot Normal \\
         PITT DeepONet\textsuperscript{*} & Burgers & 64 & 1 & 1 & [10,128,128] & [100,128,128] & silu & Glorot Normal \\
         PITT DeepONet\textsuperscript{*} & KdV & 64 & 1 & 1 & [10,128,128] & [100,128,128] & silu & Glorot Normal \\
    \end{tabular}}
    \caption{DeepONet Hyperparameters for 1D Next-Step Experiments}
    \label{tab:deeponet_ns_1d_model_hyperparams}
\end{table}

\subsection{1D Fixed-Future Training Details}
\label{app:1d_ff_params}
\begin{table}[H]
    \centering
    \resizebox{\linewidth}{!}{
    \begin{tabular}{c|c|cccccc}
         Model & Data Set & Batch Size & Learning Rate & Weight Decay & Dropout & Scheduler Step & Scheduler $\gamma$ \\
         \hline
         FNO & Heat & 64 & 1E-2 & 1E-7 & 0 & 100 & 0.5 \\
         FNO & Burgers & 64 & 1E-2 & 1E-7 & 0 & 100 & 0.5 \\
         FNO & KdV & 64 & 1E-2 & 1E-7 & 0 & 100 & 0.5 \\
         FNO & Combined & 64 & 1E-2 & 1E-7 & 0 & 100 & 0.5 \\
         PITT FNO\textsuperscript{$\dagger$} & Heat & 32 & 1E-3 & 1E-5 & 0.3 & N/A & N/A \\
         PITT FNO\textsuperscript{$\dagger$} & Burgers & 32 & 1E-3 & 1E-5 & 0.3 & N/A & N/A \\
         PITT FNO\textsuperscript{$\dagger$} & KdV & 32 & 1E-3 & 1E-5 & 0.3 & N/A & N/A \\
         PITT FNO\textsuperscript{$\dagger$} & Combined & 32 & 1E-3 & 1E-6 & 0.3 & N/A & N/A \\
         PITT FNO\textsuperscript{*} & Heat & 32 & 1E-3 & 1E-4 & 0.3 & N/A & N/A \\
         PITT FNO\textsuperscript{*} & Burgers & 32 & 1E-3 & 1E-4 & 0.3 & N/A & N/A \\
         PITT FNO\textsuperscript{*} & KdV & 32 & 1E-3 & 1E-4 & 0.3 & N/A & N/A \\
         PITT FNO\textsuperscript{*} & Combined & 32 & 1E-3 & 1E-5 & 0.3 & N/A & N/A \\
         OFormer & Heat & 32 & 1E-3 & 1E-5 & 0.2 & N/A & N/A \\
         OFormer & Burgers &  32 & 1E-3 & 1E-5 & 0.2 & N/A & N/A \\
         OFormer & KdV &  32 & 1E-3 & 1E-5 & 0.2 & N/A & N/A \\
         OFormer & Combined &  32 & 1E-3 & 1E-5 & 0.2 & N/A & N/A \\
         PITT OFormer\textsuperscript{$\dagger$} & Heat & 32 & 1E-3 & 1E-5 & 0.4 & N/A & N/A \\
         PITT OFormer\textsuperscript{$\dagger$} & Burgers & 32 & 1E-3 & 1E-5 & 0.4 & N/A & N/A \\
         PITT OFormer\textsuperscript{$\dagger$} & KdV & 32 & 1E-3 & 1E-5 & 0.4 & N/A & N/A \\
         PITT OFormer\textsuperscript{$\dagger$} & Combined & 32 & 1E-3 & 1E-6 & 0.2 & N/A & N/A \\
         PITT OFormer\textsuperscript{*} & Heat & 32 & 1E-3 & 1E-5 & 0.4 & N/A & N/A \\
         PITT OFormer\textsuperscript{*} & Burgers & 32 & 1E-3 & 1E-5 & 0.4 & N/A & N/A \\
         PITT OFormer\textsuperscript{*} & KdV & 32 & 1E-3 & 1E-5 & 0.4 & N/A & N/A \\
         PITT OFormer\textsuperscript{*} & Combined & 32 & 1E-3 & 1E-6 & 0.2 & N/A & N/A \\
         DeepONet & Heat & 32 & 1E-3 & 1e-6 & 0.01 & 200 & 0.5 \\
         DeepONet & Burgers & 32 & 1E-3 & 1e-6 & 0.01 & 200 & 0.5 \\
         DeepONet & KdV & 32 & 1E-3 & 1e-6 & 0.01 & 200 & 0.5 \\
         DeepONet & Combined & 32 & 1E-3 & 1e-6 & 0.01 & 200 & 0.5 \\
         PITT DeepONet\textsuperscript{$\dagger$} & Heat & 32 & 1E-3 & 1E-7 & 0.2 & N/A & N/A \\
         PITT DeepONet\textsuperscript{$\dagger$} & Burgers & 32 & 1E-3 & 1E-7 & 0.2 & N/A & N/A \\
         PITT DeepONet\textsuperscript{$\dagger$} & KdV & 32 & 1E-3 & 1E-4 & 0.3 & N/A & N/A \\
         PITT DeepONet\textsuperscript{$\dagger$} & Combined & 16 & 1E-3 & 1E-8 & 0.1 & N/A & N/A \\
         PITT DeepONet\textsuperscript{*} & Heat & 16 & 1E-4 & 1E-5 & 0.2 & N/A & N/A \\
         PITT DeepONet\textsuperscript{*} & Burgers & 16 & 1E-4 & 1E-5 & 0.2 & N/A & N/A \\
         PITT DeepONet\textsuperscript{*} & KdV & 16 & 1E-4 & 1E-5 & 0.2 & N/A & N/A \\
         PITT DeepONet\textsuperscript{*} & Combined & 16 & 1E-4 & 1E-7 & 0.2 & N/A & N/A \\
    \end{tabular}}
    \caption{Training Hyperparameters for 1D Fixed-Future Experiments}
\end{table}

\begin{table}[H]
    \centering
    \resizebox{\linewidth}{!}{
    \begin{tabular}{c|c|ccccccc}
         Model & Data Set & Hidden Dimension & Numerical Hidden Dimension & Numerical Layers & Heads & FNO Modes \\
         \hline
         FNO & Heat & 256 & N/A & N/A & N/A & 8 \\
         FNO & Burgers & 256 & N/A & N/A & N/A & 8 \\
         FNO & KdV & 256 & N/A & N/A & N/A & 8 \\
         FNO & Combined & 256 & N/A & N/A & N/A & 8 \\
         PITT FNO\textsuperscript{$\dagger$} & Heat & 64 & 32 & 2 & 2 & 6 \\
         PITT FNO\textsuperscript{$\dagger$} & Burgers &  64 & 32 & 2 & 2 & 6 \\
         PITT FNO\textsuperscript{$\dagger$} & KdV &  64 & 32 & 2 & 2 & 6 \\
         PITT FNO\textsuperscript{$\dagger$} & Combined &  64 & 32 & 2 & 2 & 6 \\
         PITT FNO\textsuperscript{*} & Heat &  64 & 32 & 2 & 2 & 6 \\
         PITT FNO\textsuperscript{*} & Burgers &  64 & 32 & 2 & 2 & 6 \\
         PITT FNO\textsuperscript{*} & KdV &  64 & 32 & 2 & 2 & 6 \\
         PITT FNO\textsuperscript{*} & Combined &  64 & 32 & 2 & 2 & 6 \\
    \end{tabular}}
    \caption{FNO Hyperparameters for 1D Fixed-Future Experiments}
\end{table}
\begin{table}[H]
    \centering
    \resizebox{\linewidth}{!}{
    \begin{tabular}{c|c|ccccccccccc}
         Model & Data Set & Hidden Dim. & Numerical Layers & Heads & Input Embedding Dim. & Output Embedding Dim. & Encoder Depth & Decoder Depth & Latent Channels & Encoder Resolution & Decoder Resolution  & Scale\\
         \hline
         OFormer & Heat & N/A & N/A & N/A & 64 & 256 & 2 & 2 & 256 & 1024 & 1024 & 8 \\
         OFormer & Burgers & N/A & N/A & N/A &  64 & 256 & 2 & 2 & 256 & 1024 & 1024 & 8 \\
         OFormer & KdV & N/A & N/A & N/A &  64 & 256 & 2 & 2 & 256 & 1024 & 1024 & 8 \\
         OFormer & Combined & N/A & N/A & N/A &  64 & 256 & 2 & 2 & 256 & 1024 & 1024 & 8 \\
         PITT OFormer\textsuperscript{$\dagger$} & Heat & 16 & 4 & 4 & 32 & 32 & 2 & 2 & 32 & 32 & 32 & 8 \\
         PITT OFormer\textsuperscript{$\dagger$} & Burgers & 16 & 4 & 4 & 32 & 32 & 2 & 2 & 32 & 32 & 32 & 8 \\
         PITT OFormer\textsuperscript{$\dagger$} & KdV & 16 & 4 & 4 & 32 & 32 & 2 & 2 & 32 & 32 & 32 & 8 \\
         PITT OFormer\textsuperscript{$\dagger$} & Combined & 16 & 4 & 4 & 32 & 32 & 2 & 2 & 32 & 32 & 32 & 8 \\
         PITT OFormer\textsuperscript{*} & Heat & 16 & 4 & 4 & 32 & 32 & 2 & 2 & 32 & 32 & 32 & 8 \\
         PITT OFormer\textsuperscript{*} & Burgers & 16 & 4 & 4 & 32 & 32 & 2 & 2 & 32 & 32 & 32 & 8 \\
         PITT OFormer\textsuperscript{*} & KdV & 16 & 4 & 4 & 32 & 32 & 2 & 2 & 32 & 32 & 32 & 8 \\
         PITT OFormer\textsuperscript{*} & Combined & 16 & 4 & 4 & 32 & 32 & 2 & 2 & 32 & 32 & 32 & 8 \\
    \end{tabular}}
    \caption{OFormer Hyperparameters for 1D Fixed-Future Experiments}
\end{table}
\begin{table}[H]
    \centering
    \resizebox{\linewidth}{!}{
    \begin{tabular}{c|c|cccccccc}
         Model & Data Set & Hidden Dimension & Numerical Layers & Heads & Branch Net & Trunk Net & Activation & Initializer \\
         \hline
         DeepONet & Heat & N/A & N/A & N/A & [10, 256, 256] & [100, 256, 256] & SiLU & Glorot Normal \\
         DeepONet & Burgers & N/A & N/A & N/A & [10, 256, 256] & [100, 256, 256] & SiLU & Glorot Normal \\
         DeepONet & KdV & N/A & N/A & N/A & [10, 256, 256] & [100, 256, 256] & SiLU & Glorot Normal \\
         DeepONet & Combined & N/A & N/A & N/A & [10, 256, 256] & [100, 256, 256] & SiLU & Glorot Normal \\
         PITT DeepONet\textsuperscript{$\dagger$} & Heat & 32 & 2 & 2 & [10, 256, 256] & [100, 256, 256] & SiLU & Glorot Normal \\
         PITT DeepONet\textsuperscript{$\dagger$} & Burgers & 32 & 2 & 2 & [10, 256, 256] & [100, 256, 256] & SiLU & Glorot Normal \\
         PITT DeepONet\textsuperscript{$\dagger$} & KdV & 32 & 2 & 2 & [10, 256, 256] & [100, 256, 256] & SiLU & Glorot Normal \\
         PITT DeepONet\textsuperscript{$\dagger$} & Combined & 32 & 2 & 2 & [10, 256, 256] & [100, 256, 256] & SiLU & Glorot Normal \\
         PITT DeepONet\textsuperscript{*} & Heat & 32 & 2 & 2 & [10, 256, 256] & [100, 256, 256] & SiLU & Glorot Normal \\
         PITT DeepONet\textsuperscript{*} & Burgers & 32 & 2 & 2 & [10, 256, 256] & [100, 256, 256] & SiLU & Glorot Normal \\
         PITT DeepONet\textsuperscript{*} & KdV & 32 & 2 & 2 & [10, 256, 256] & [100, 256, 256] & SiLU & Glorot Normal \\
         PITT DeepONet\textsuperscript{*} & Combined & 32 & 2 & 2 & [10, 256, 256] & [100, 256, 256] & SiLU & Glorot Normal \\
    \end{tabular}}
    \caption{DeepONet Hyperparameters for 1D Fixed-Future Experiments}
    \label{tab:deeponet_ff_1d_hyperparameters}
\end{table}

\subsection{2D Navier-Stokes Next-Step Training Details}
\label{app:2d_ns_ns_params}

All models were trained for 100 epochs on the 2D Navier-Stokes data for next-step training and rollout testing.
Note: the heads hyperparameter controls the number of heads for both the self attention and linear attention blocks.

\begin{table}[H]
    \centering
    \resizebox{\linewidth}{!}{
    \begin{tabular}{c|cccccc}
         Model & Batch Size & Learning Rate & Weight Decay & Dropout & Scheduler Step & Scheduler $\gamma$ \\ 
         \hline
         FNO & 8 & 1E-4 & 1E-5 & 0 & 10 & 0.5\\
         PITT FNO\textsuperscript{$\dagger$} & 8 & 1E-4 & 0 & 0 & N/A & N/A \\
         PITT FNO\textsuperscript{*} & 64 & 1E-3 & 1E-7 & 0 & N/A & N/A \\
         OFormer & 8 & 1E-3 & 0 & 0 & N/A & N/A \\
         PITT OFormer\textsuperscript{$\dagger$} & 16 & 1E-3 & 0 & 0 & N/A & N/A \\
         PITT OFormer\textsuperscript{*} & 8 & 1E-4 & 1E-6 & 0 & N/A & N/A \\
         DeepONet & 8 & 1E-4 & 1E-7 & 0 & 20 & 0.5 \\
         PITT DeepONet\textsuperscript{$\dagger$} & 4 & 1E-4 & 0 & 0 & N/A & N/A \\
         PITT DeepONet\textsuperscript{*} & 16 & 1E-4 & 0 & 0 & N/A & N/A \\
    \end{tabular}}
    \caption{Training Hyperparameters for the 2D Navier-Stokes Next-Step Experiment}
    \label{tab:pitt_2d_ns_ns_train_hyperparams}
\end{table}

\begin{table}[H]
    \centering
    \resizebox{\linewidth}{!}{
    \begin{tabular}{c|cccccc}
         Model & Hidden Dimension & Numerical Layers & Heads & FNO Modes 1 & FNO Modes 2\\
         \hline
         FNO & 64 & N/A & N/A & 8 & 8 \\
         PITT FNO\textsuperscript{$\dagger$} & 32 & 8 & 4 & 8 & 8 \\
         PITT FNO\textsuperscript{*} & 32 & 8 & 4 & 8 & 8 \\
    \end{tabular}}
    \caption{Model Hyperparameters for the 2D Navier-Stokes Next-Step Experiment}
    \label{tab:pitt_2d_ns_ns_model_hyperparams}
\end{table}
\begin{table}[H]
    \centering
    \resizebox{\linewidth}{!}{
    \begin{tabular}{c|ccccccccccc}
         Model & Hidden Dim. & Numerical Layers & Heads & Input Embedding Dim. & Output Embedding Dim. & Encoder Depth & Decoder Depth & Latent Channels & Encoder Resolution & Decoder Resolution  & Scale\\
         \hline
         OFormer & N/A & N/A & 4 & 128 & 128 & 2 & 1 & 128 & 128 & 128 & 16 \\
         PITT OFormer\textsuperscript{$\dagger$} & 32 & 1 & 4 & 64 & 64 & 2 & 1 & 64 & 64 & 64 & 16 \\
         PITT OFormer\textsuperscript{*} & 64 & 2 & 2 & 64 & 64 & 2 & 1 & 64 & 64 & 128 & 16\\
    \end{tabular}}
    \caption{Model Hyperparameters for the 2D Navier-Stokes Next-Step Experiment}
    \label{tab:pitt_2d_ns_ns_model_hyperparams}
\end{table}
\begin{table}[H]
    \centering
    \resizebox{\linewidth}{!}{
    \begin{tabular}{c|cccccccc}
         Model & Hidden Dimension & Numerical Layers & Heads & Branch Net & Trunk Net & Activation & Initializer \\
         \hline
         DeepONet & N/A & N/A & N/A & [1,256,256,256] & [2,256,256,256] & relu & Glorot Normal \\
         PITT DeepONet\textsuperscript{$\dagger$} & 32 & 5 & 4 & [1,128,128] & [2,128,128] & silu & Glorot Normal \\
         PITT DeepONet\textsuperscript{*} & 16 & 20 & 4 & [1,128,128] & [2,128,218] & silu & Glorot Normal\\
    \end{tabular}}
    \caption{Model Hyperparameters for the 2D Navier-Stokes Next-Step Experiment}
    \label{tab:pitt_2d_ns_ns_model_hyperparams}
\end{table}

\subsection{2D Navier-Stokes Fixed-Future Training Details}
\label{app:2d_ns_ff_params}
All models were trained for 200 epochs on the 2D Navier-Stokes data for the fixed-future experiments.
\begin{table}[H]
    \centering
    \resizebox{\linewidth}{!}{
    \begin{tabular}{c|cccccc}
         Model & Batch Size & Learning Rate & Weight Decay & Dropout & Scheduler Step & Scheduler $\gamma$ \\ 
         \hline
         FNO $T=20s$ & 8 & $1E-3$ & $1E-5$ & 0.0 & 40 & 0.5 \\
         FNO $T=30s$ & 8 & $1E-3$ & $1E-5$ & 0.0 & 40 & 0.5 \\
         PITT FNO\textsuperscript{$\dagger$} $T=20s$ & 16 & $1E-2$ & $1E-5$ & 0.0 & N/A & N/A \\
         PITT FNO\textsuperscript{$\dagger$} $T=30s$ & 16 & $1E-2$ & $1E-5$ & 0.0 & N/A & N/A \\
         PITT FNO\textsuperscript{*} $T=20s$ & 16 & $1E-2$ & $1E-5$ & 0.0 & N/A & N/A \\
         PITT FNO\textsuperscript{*} $T=30s$ & 16 & $1E-2$ & $1E-5$ & 0.0 & N/A & N/A \\
         OFormer $T=20s$ & 8 & 1E-3 & 1E-8 & 0.1 & N/A & N/A \\
         OFormer $T=30s$ & 8 & 1E-4 & 1E-8 & 0.1 & N/A & N/A \\
         PITT OFormer\textsuperscript{$\dagger$} $T=20s$ & 8 & 1E-3 & 1E-8 & 0.1 & N/A & N/A \\
         PITT OFormer\textsuperscript{$\dagger$} $T=30s$ & 16 & 1E-3 & 1E-7 & 0.1 & N/A & N/A \\
         PITT OFormer\textsuperscript{*} $T=20s$ & 8 & 1E-3 & 1E-4 & 0.5 & N/A & N/A \\
         PITT OFormer\textsuperscript{*} $T=30s$ & 16 & 1E-3 & 1E-4 & 0.6 & N/A & N/A \\
         DeepONet $T=20s$ & 32 & 1E-3 & 1E-6 & 0.0 & 20 & 0.5 \\
         DeepONet $T=30s$ & 32 & 1E-3 & 1E-7 & 0.0 & 20 & 0.5 \\
         PITT DeepONet\textsuperscript{$\dagger$} $T=20s$ & 16 & 1E-3 & 1E-6 & 0.2 & N/A & N/A \\
         PITT DeepONet\textsuperscript{$\dagger$} $T=30s$ & 16 & 1E-3 & 1E-7 & 0.2 & N/A & N/A \\
         PITT DeepONet\textsuperscript{*} $T=20s$ & 8 & 1E-4 & 0.0 & 0.0 & N/A & N/A \\
         PITT DeepONet\textsuperscript{*} $T=30s$ & 8 & 1E-4 & 0.0 & 0.0 & N/A & N/A \\
    \end{tabular}}
    \caption{Training Hyperparameters for the 2D Navier-Stokes Fixed-Future Experiment}
    \label{tab:pitt_2d_ns_ff_train_hyperparams}
\end{table}

\begin{table}[H]
    \centering
    \resizebox{\linewidth}{!}{
    \begin{tabular}{c|ccccccc}
         Model & FNO Hidden Dimension & Transformer Hidden Dimension & Numerical Layers & Heads & FNO Modes 1 & FNO Modes 2\\
         \hline
         FNO $T=20s$ & 32 & N/A & N/A N/A & N/A & 6 & 6 \\
         FNO $T=30s$ & 32 & N/A & N/A & N/A & 6 & 6 \\
         PITT FNO\textsuperscript{$\dagger$} $T=20s$ & 32 & 16 & 20 & 4 & 4 & 4 \\
         PITT FNO\textsuperscript{$\dagger$} $T=30s$ & 32 & 16 & 20 & 4 & 4 & 4 \\
         PITT FNO\textsuperscript{*} $T=20s$ & 32 & 16 & 20 & 4 & 4 & 4 \\
         PITT FNO\textsuperscript{*} $T=30s$ & 32 & 16 & 20 & 4 & 4 & 4 \\
    \end{tabular}}
    \caption{Model Hyperparameters for the 2D Navier-Stokes Fixed-Future Experiment}
    \label{tab:pitt_2d_ns_ff_model_hyperparams}
\end{table}
\begin{table}[H]
    \centering
    \resizebox{\linewidth}{!}{
    \begin{tabular}{c|ccccccccccc}
         Model & Hidden Dim. & Numerical Layers & Heads & Input Embedding Dim. & Output Embedding Dim. & Encoder Depth & Decoder Depth & Latent Channels & Encoder Resolution & Decoder Resolution  & Scale\\
         \hline
         OFormer $T=20s$ & N/A & N/A & 4 & 64 & 64 & 2 & 1 & 64 & 64 & 128 & 16 \\
         OFormer $T=30s$ & N/A & N/A & 4 & 64 & 64 & 2 & 1 & 64 & 64 & 128 & 16 \\
         PITT OFormer\textsuperscript{$\dagger$} $T=20s$ & 32 & 5 & 2 & 64 & 64 & 2 & 1 & 64 & 64 & 64 & 16 \\
         PITT OFormer\textsuperscript{$\dagger$} $T=30s$ & 32 & 5 & 2 & 64 & 64 & 2 & 1 & 64 & 64 & 64 & 16 \\
         PITT OFormer\textsuperscript{*} $T=20s$ & 64 & 4 & 4 & 64 & 64 & 2 & 1 & 64 & 64 & 128 & 16\\
         PITT OFormer\textsuperscript{*} $T=30s$ & 64 & 4 & 4 & 64 & 64 & 2 & 1 & 64 & 64 & 128 & 16\\
    \end{tabular}}
    \caption{Model Hyperparameters for the 2D Navier-Stokes Fixed-Future Experiment}
    \label{tab:pitt_2d_ns_ff_model_hyperparams}
\end{table}
\begin{table}[H]
    \centering
    \resizebox{\linewidth}{!}{
    \begin{tabular}{c|cccccccc}
         Model & Hidden Dimension & Numerical Layers & Heads & Branch Net & Trunk Net & Activation & Initializer \\
         \hline
         DeepONet  $T=20s$ & N/A & N/A & N/A & [41,256,256,256] & [2,256,256,256] & relu & Glorot Normal \\
         DeepONet  $T=30s$ & N/A & N/A & N/A & [41,256,256,256] & [2,256,256,256] & relu & Glorot Normal \\
         PITT DeepONet\textsuperscript{$\dagger$} $T=20s$ & 32 & 10 & 8 & [41,128,128] & [2,128,128] & silu & Glorot Normal \\
         PITT DeepONet\textsuperscript{$\dagger$} $T=30s$ & 32 & 10 & 8 & [41,128,128] & [2,128,128] & silu & Glorot Normal \\
         PITT DeepONet\textsuperscript{*} $T=20s$ & 32 & 5 & 2 & [41,128,128] & [2,128,128] & silu & Glorot Normal \\
         PITT DeepONet\textsuperscript{*} $T=30s$ & 32 & 5 & 2 & [41,128,128] & [2,128,128] & silu & Glorot Normal \\
    \end{tabular}}
    \caption{Model Hyperparameters for the 2D Navier-Stokes Fixed-Future Experiment}
    \label{tab:pitt_2d_ns_ff_model_hyperparams}
\end{table}

\subsection{2D Poisson Steady-State Training Details}
\label{app:2d_ns_poisson_params}
All models were trained for 1000 epochs on the 2D Poisson steady-state data.
\begin{table}[H]
    \centering
    \resizebox{\linewidth}{!}{
    \begin{tabular}{c|cccccc}
         Model & Batch Size & Learning Rate & Weight Decay & Dropout & Scheduler Step & Scheduler $\gamma$ \\ 
         \hline
         FNO & 128 & 1E-3 & 1E-7 & 0.1 & 200 & 0.5 \\
         PITT FNO\textsuperscript{$\dagger$} & 128 & 1E-3 & 0 & 0.05 & N/A & N/A \\
         PITT FNO\textsuperscript{*} & 256 & 1E-3 & 0 & 0.05 & N/A & N/A \\
         OFormer & 32  & 1E-4 & 1E-8 & 0 & N/A & N/A \\
         PITT OFormer\textsuperscript{$\dagger$} & 64 & 1E-3 & 0 & 0 & N/A & N/A \\
         PITT OFormer\textsuperscript{*} & 128 & 1E-3 & 0 & 0 & N/A & N/A \\
         DeepONet & 128 & 1E-3 & 1E-8 & 0.0 & 100 & 0.5 \\
         PITT DeepONet\textsuperscript{$\dagger$} & 32 & 1E-3 & 1E-8 & 0 & N/A & N/A \\
         PITT DeepONet\textsuperscript{*} & 256 & 1E-3 & 0 & 0 & N/A & N/A \\
    \end{tabular}}
    \caption{Training Hyperparameters for the 2D Poisson Experiment}
    \label{tab:pitt_2d_poisson_train_hyperparams}
\end{table}

\begin{table}[H]
    \centering
    \resizebox{\linewidth}{!}{
    \begin{tabular}{c|ccccccc}
         Model & FNO Hidden Dimension & Transformer Hidden Dimension & Numerical Layers & Heads & FNO Modes 1 & FNO Modes 2\\
         \hline
         FNO & 128 & N/A & N/A & N/A & 8 & 8 \\
         PITT FNO\textsuperscript{$\dagger$} & 64 & 64 & 8 & 8 & 8 & 8 \\
         PITT FNO\textsuperscript{*} & 64 & 64 & 8 & 8 & 8 & 8 \\
    \end{tabular}}
    \caption{Model Hyperparameters for the 2D Poisson Experiment}
    \label{tab:pitt_2d_poisson_model_hyperparams}
\end{table}
\begin{table}[H]
    \centering
    \resizebox{\linewidth}{!}{
    \begin{tabular}{c|ccccccccccc}
         Model & Hidden Dim. & Numerical Layers & Heads & Input Embedding Dim. & Output Embedding Dim. & Encoder Depth & Decoder Depth & Latent Channels & Encoder Resolution & Decoder Resolution  & Scale\\
         \hline
         OFormer & N/A & N/A & 2 & 128 & 128 & 2 & 1 & 128 & 128 & 128 & 16 \\
         PITT OFormer\textsuperscript{$\dagger$} & 64 & 10 & 4 & 64 & 64 & 2 & 1 & 64 & 64 & 64 & 16 \\
         PITT OFormer\textsuperscript{*} & 64 & 10 & 4 & 64 & 64 & 2 & 1 & 64 & 64 & 64 & 16 \\
    \end{tabular}}
    \caption{Model Hyperparameters for the 2D Poisson Experiment}
    \label{tab:pitt_2d_poisson_model_hyperparams}
\end{table}
\begin{table}[H]
    \centering
    \resizebox{\linewidth}{!}{
    \begin{tabular}{c|cccccccc}
         Model & Hidden Dimension & Numerical Layers & Heads & Branch Net & Trunk Net & Activation & Initializer \\
         \hline
         DeepONet & N/A & N/A & N/A & [1,256,256,256,256] & [2,256,256,256,256] & relu & Glorot Normal \\
         PITT DeepONet\textsuperscript{*} & 64 & 8 & 8 & [1,512,512,512] & [2,512,512,512] & relu & Glorot Normal \\
         PITT DeepONet\textsuperscript{$\dagger$} & 64 & 10 & 4 & [1,256,256] & [2,256,256] & silu & Glorot Normal \\
    \end{tabular}}
    \caption{Model Hyperparameters for the 2D Poisson Experiment}
    \label{tab:pitt_2d_poisson_model_hyperparams}
\end{table}

\newpage
\section{1D Standard Embedding Output Decomposition}
\begin{figure}[ht]
    \centering
    \includegraphics[width=0.95\linewidth]{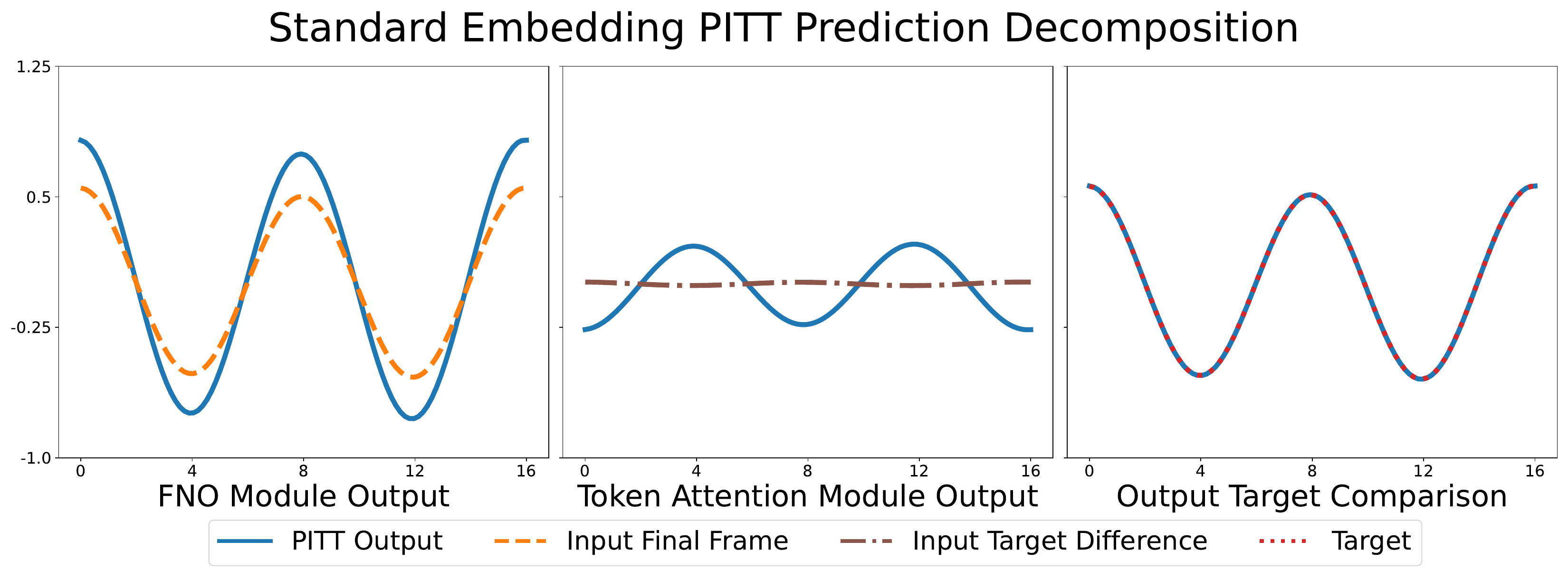}
    \caption{PITT FNO prediction decomposition for 1D Heat equation. \textbf{Left:} The FNO module of PITT predicts a large change to the final frame of input data. \textbf{Middle} The numerical update block corrects the FNO output. \textbf{Right} The combination of FNO and numerical update block output very accurately predicts the next step.}
    \label{fig:pitt_decomp_standard}
\end{figure}

\newpage
\section{PITT Attention Maps}
\begin{figure}
    \centering
    \begin{subfigure}[b]{\linewidth}
        \caption{$\quad\quad\quad\quad\quad\quad\quad\quad\quad\quad\quad\quad\quad\quad\quad\quad\quad\quad\quad\quad\quad\quad\quad\quad\quad\quad\quad\quad\quad\quad\quad\quad\quad\quad\quad\quad\quad\quad\quad\quad\quad$}
        \includegraphics[width=0.95\linewidth]{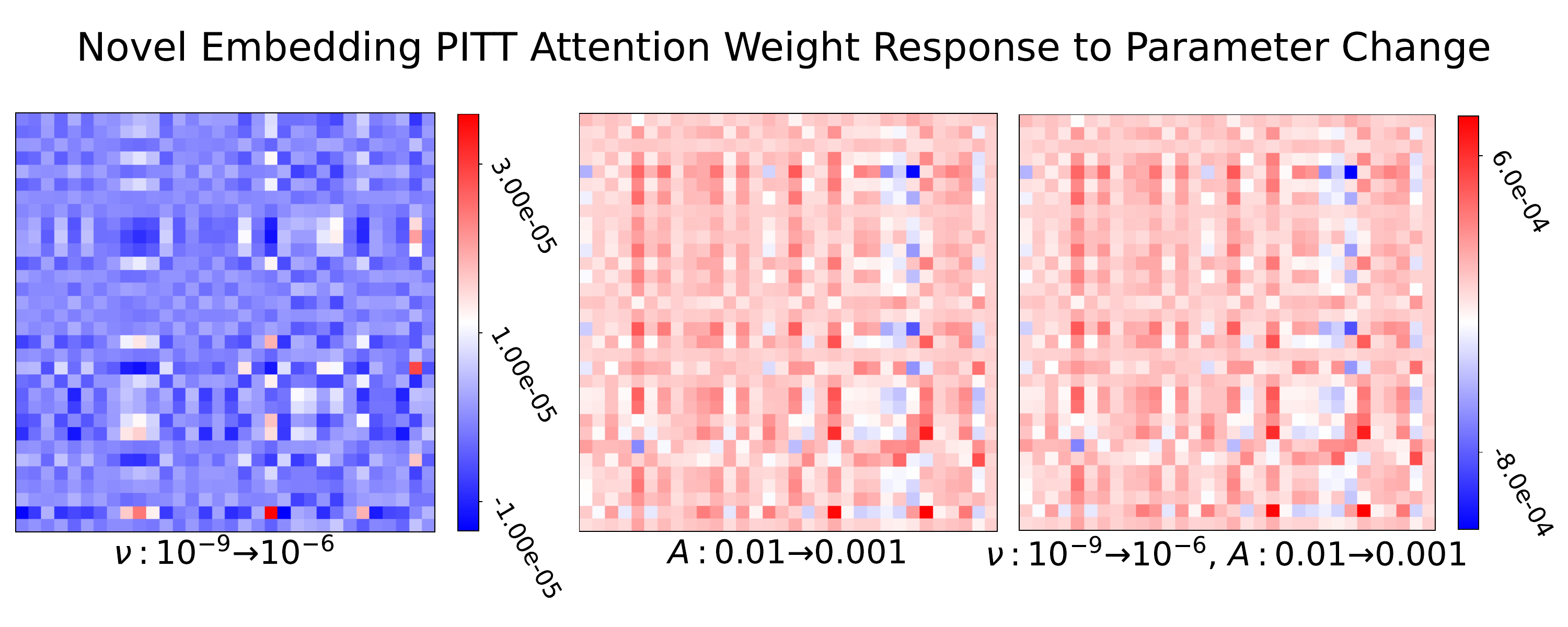}
        \label{fig:old_pitt_token_exploration}
    \end{subfigure}
    \begin{subfigure}[b]{\linewidth}
        \caption{$\quad\quad\quad\quad\quad\quad\quad\quad\quad\quad\quad\quad\quad\quad\quad\quad\quad\quad\quad\quad\quad\quad\quad\quad\quad\quad\quad\quad\quad\quad\quad\quad\quad\quad\quad\quad\quad\quad\quad\quad\quad$}
        \includegraphics[width=0.95\linewidth]{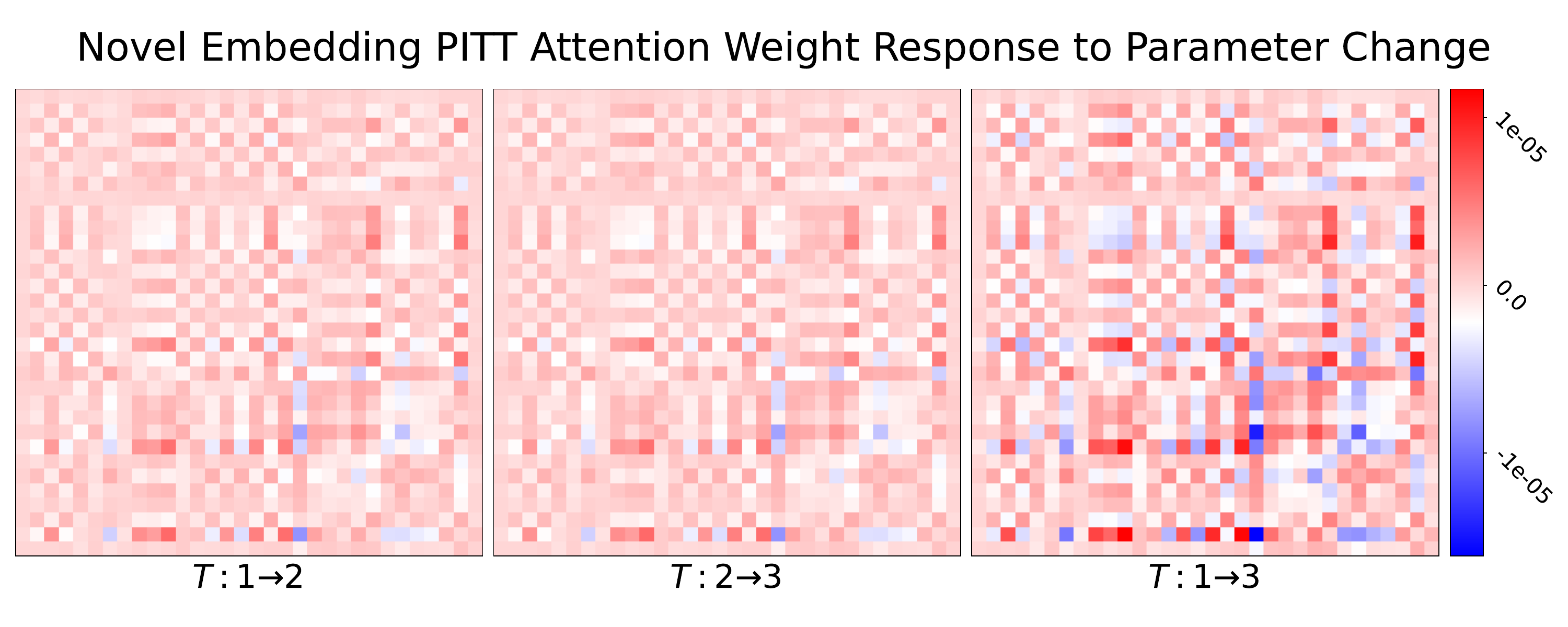}
        \label{fig:old_pitt_token_exploration_time}
    \end{subfigure}
    \caption{PITT attention weight response to changing equation tokens. \textbf{a)} PITT attention weights change as we modify the input tokens. From our Navier-Stokes equation, we see the attention weights change differently when we modify the viscosity and forcing term amplitude.
        This demonstrates that PITT is able to learn equation parameters from the tokenized equations.
        \textbf{b)} PITT attention weights change as we modify the input token target time.
        From our Navier-Stokes equation, we see the attention weights change differently when we modify the target time.
        This demonstrates that PITT is able to learn time evolution from the tokenized equations.}
\end{figure}

\begin{figure}
    \centering
    \begin{subfigure}[b]{\linewidth}
        \caption{$\quad\quad\quad\quad\quad\quad\quad\quad\quad\quad\quad\quad\quad\quad\quad\quad\quad\quad\quad\quad\quad\quad\quad\quad\quad\quad\quad\quad\quad\quad\quad\quad\quad\quad\quad\quad\quad\quad\quad\quad\quad$}
        \includegraphics[width=0.95\linewidth]{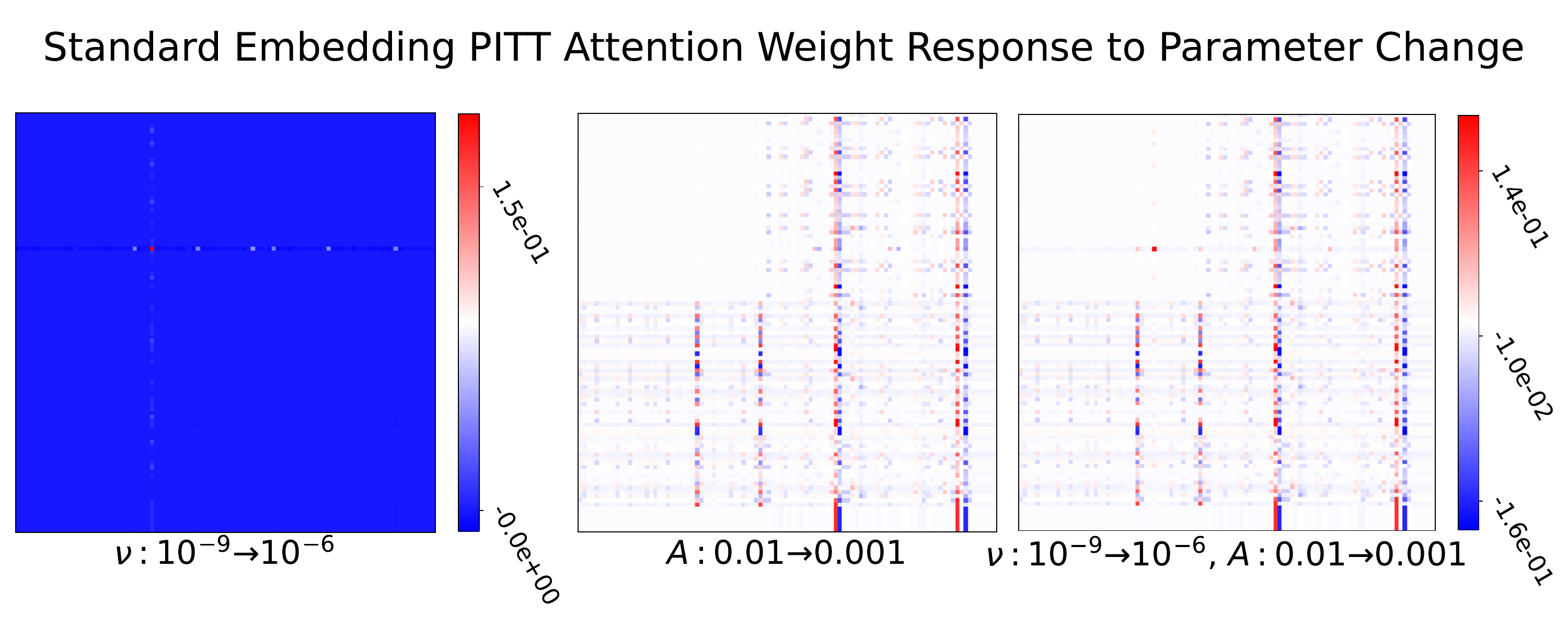}
        \label{fig:new_pitt_token_exploration}
    \end{subfigure}
    \begin{subfigure}[b]{\linewidth}
        \caption{$\quad\quad\quad\quad\quad\quad\quad\quad\quad\quad\quad\quad\quad\quad\quad\quad\quad\quad\quad\quad\quad\quad\quad\quad\quad\quad\quad\quad\quad\quad\quad\quad\quad\quad\quad\quad\quad\quad\quad\quad\quad$}
        \includegraphics[width=0.95\linewidth]{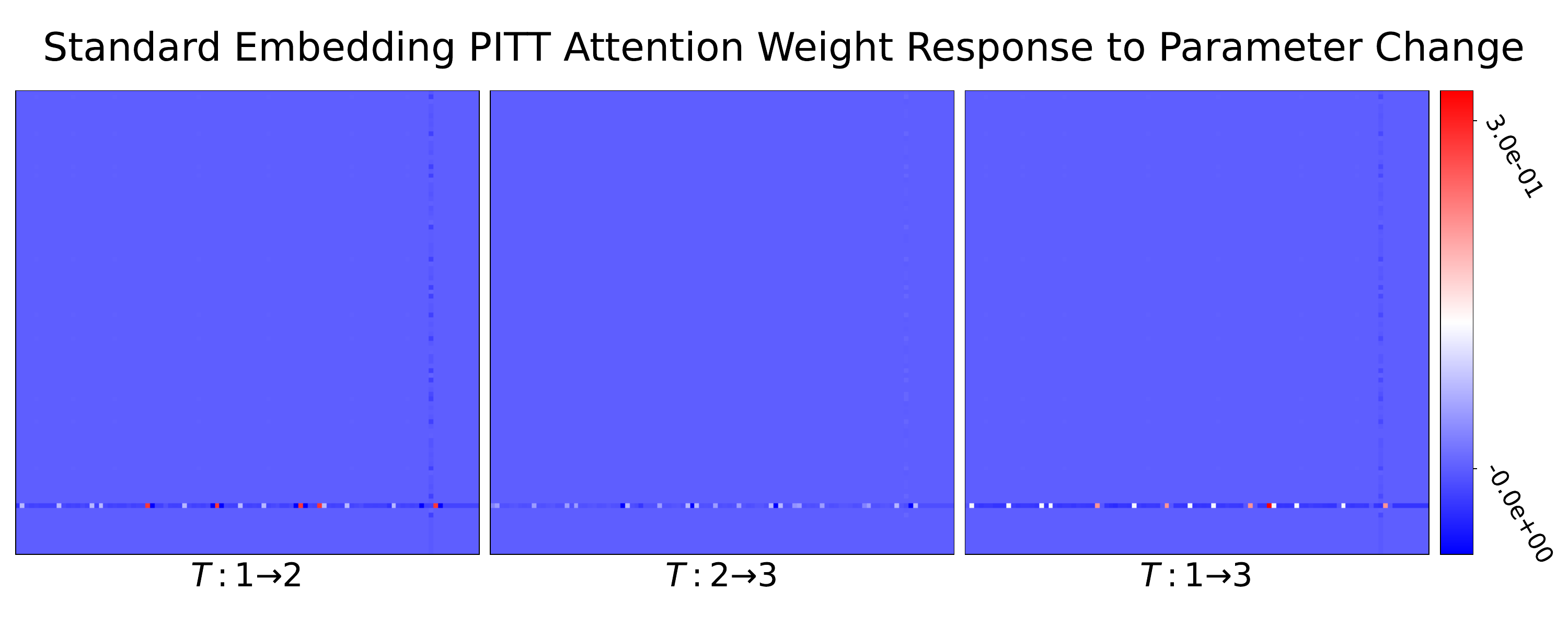}
        \label{fig:new_pitt_token_exploration_time}
    \end{subfigure}
    \caption{PITT attention weight response to changing equation tokens. \textbf{a)} PITT attention weights change as we modify the input tokens. From our Navier-Stokes equation, we see the attention weights change differently when we modify the viscosity and forcing term amplitude.
        This demonstrates that PITT is able to learn equation parameters from the tokenized equations.
        \textbf{b)} PITT attention weights change as we modify the input token target time.
        From our Navier-Stokes equation, we see the attention weights change differently when we modify the target time.
        This demonstrates that PITT is able to learn time evolution from the tokenized equations.}
\end{figure}

Having the analytical governing equations explicitly used as input allows us to easily test the effect of equation parameters on our model.
We modify the viscosity forcing term amplitude in the analytical equation and plot the difference in self attention weights between our initial and modified equations.
These attention weights come from the self-attention block seen in figure \ref{fig:token_transformer}.
Seen in below figure \ref{fig:new_pitt_token_exploration}, \ref{fig:new_pitt_token_exploration_time}, \ref{fig:old_pitt_token_exploration}, and \ref{fig:old_pitt_token_exploration_time} the attention weights from the self-attention block used for latent equation learning clearly shows distinctive behavior if the viscosity or forcing term amplitude is modified.
This is expected because those parameters control substantially different properties in our system.
The forcing term amplitude also dominates the viscosity in attention weight difference for our novel embedding, where the attention map differences are of similar magnitude for standard embedding.
Another key feature of tokenizing equations directly is that we are able to explicitly add our target evaluation time into the embedding.
In our next-step style training, this is the simulation time for the target frame.
For example, using a timestep of 0.05, if we were to use the first second of simulation data to predict the frame at time 1.05, the target time in the tokenized equation is 1.05.
We can also visualize the attention weights after incrementing the target time to determine how well PITT is able to learn time-evolution.
This is seen in figures \ref{fig:old_pitt_token_exploration_time} and \ref{fig:new_pitt_token_exploration_time}.
As we update target time, the activation pattern in attention weight differences remains approximately constant across different target times for our novel embedding.
Similarly, for standard embedding changing time from 1 to 3 seconds results in attention map that is approximately the sum of attention maps from 1 to 2 seconds and 2 to 3 seconds.

\newpage
\section{Novel Embedding Rollout Error Accumulation Plot}
\begin{figure}[H]
    \centering
    \includegraphics[width=0.95\linewidth]{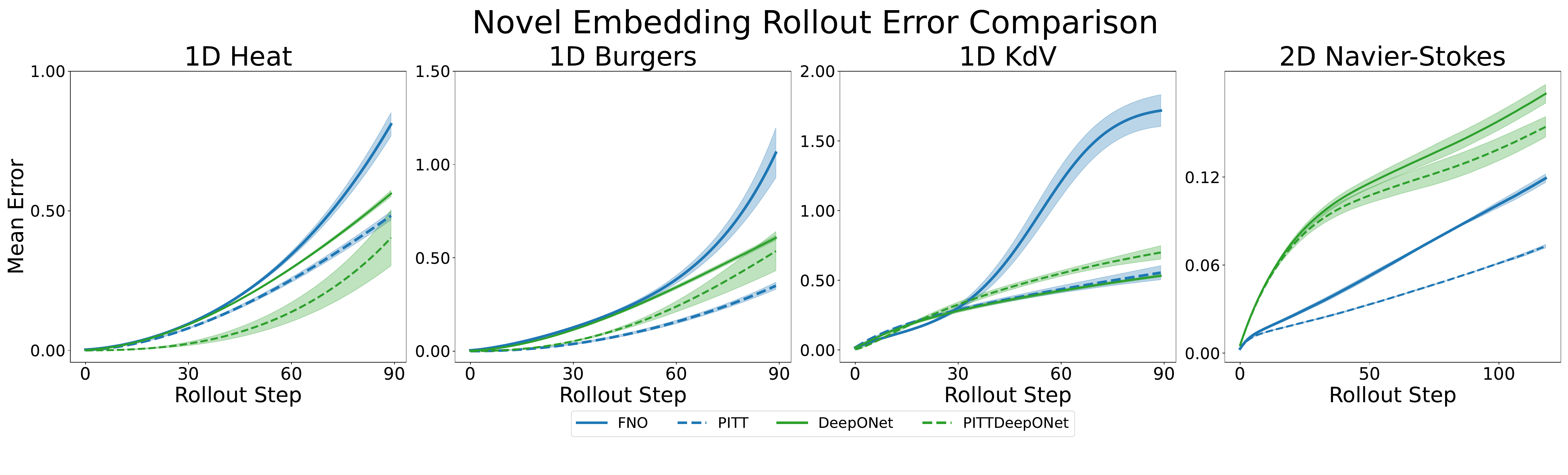}
    \caption{Error accumulation for rollout experiments. PITT variants have less error accumulation at long rollout times for every benchmark when compared to the baseline models.}
    \label{fig:1d_rollout_novel}
\end{figure}
\section{1D Rollout Comparison}
\label{app:1d_rollout}

In 1D rollout we see significantly PITT variants of FNO and DeepONet match the ground truth values much better for the Heat and Burgers simulations, and maintains its shape closer to ground truth for the KdV equation when compared to FNO.
Darker lines correspond with londer times in rollout, up to a time of 4 seconds.
\begin{figure}[H]
    \centering
    \includegraphics[width=\linewidth]{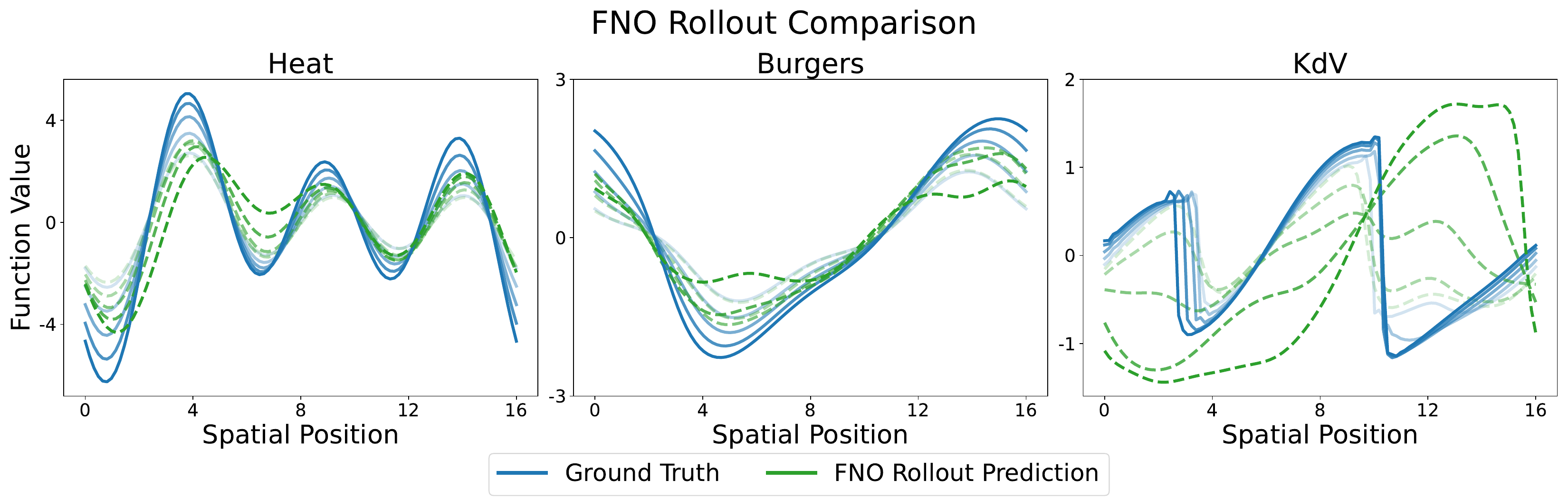}
    \caption{Comparison of FNO to ground truth data for autoregressive rollout on our 1D data sets.}
    \label{fig:pitt_1d_rollout}
\end{figure}
\begin{figure}[H]
    \centering
    \includegraphics[width=\linewidth]{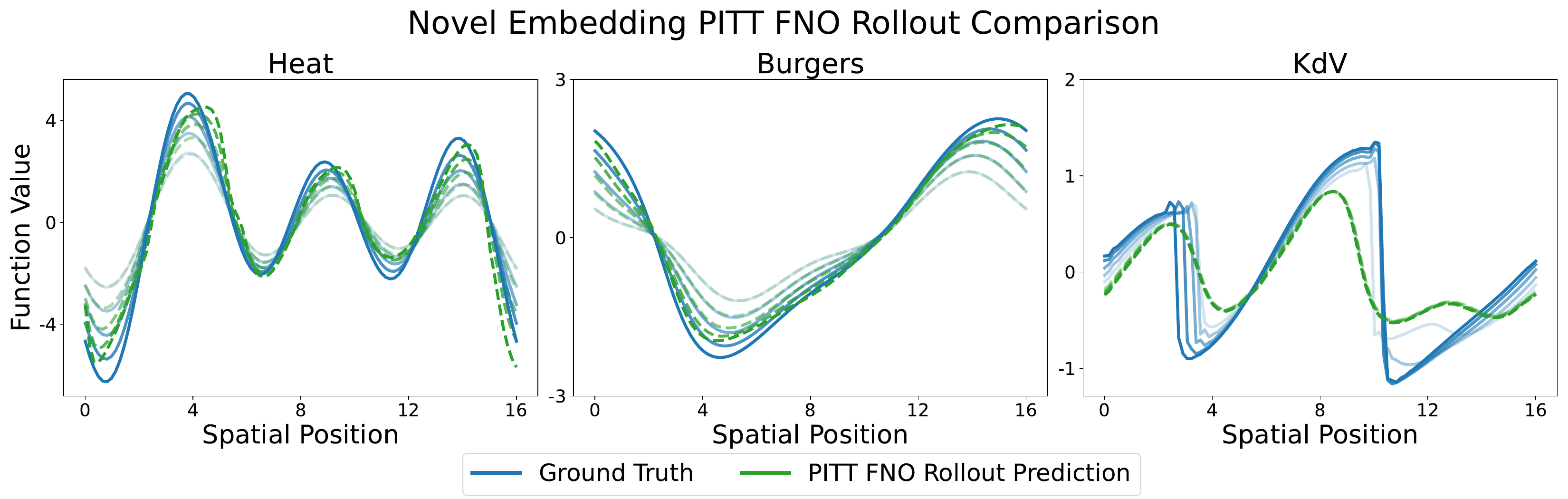}
    \caption{Comparison of PITT FNO using our novel embedding to ground truth data for autoregressive rollout on our 1D data sets.}
    \label{fig:pitt_2d_rollout}
\end{figure}
\begin{figure}[H]
    \centering
    \includegraphics[width=\linewidth]{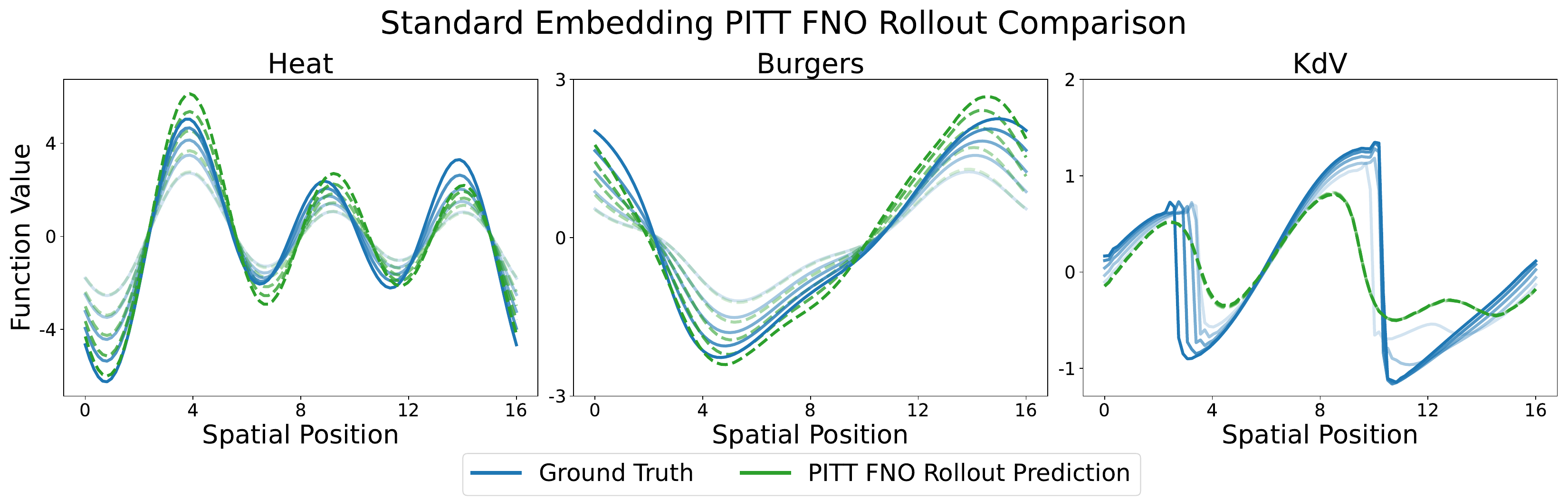}
    \caption{Comparison of PITT FNO using standard embedding to ground truth data for autoregressive rollout on our 1D data sets.}
    \label{fig:pitt_2d_rollout}
\end{figure}
\begin{figure}[H]
    \centering
    \includegraphics[width=\linewidth]{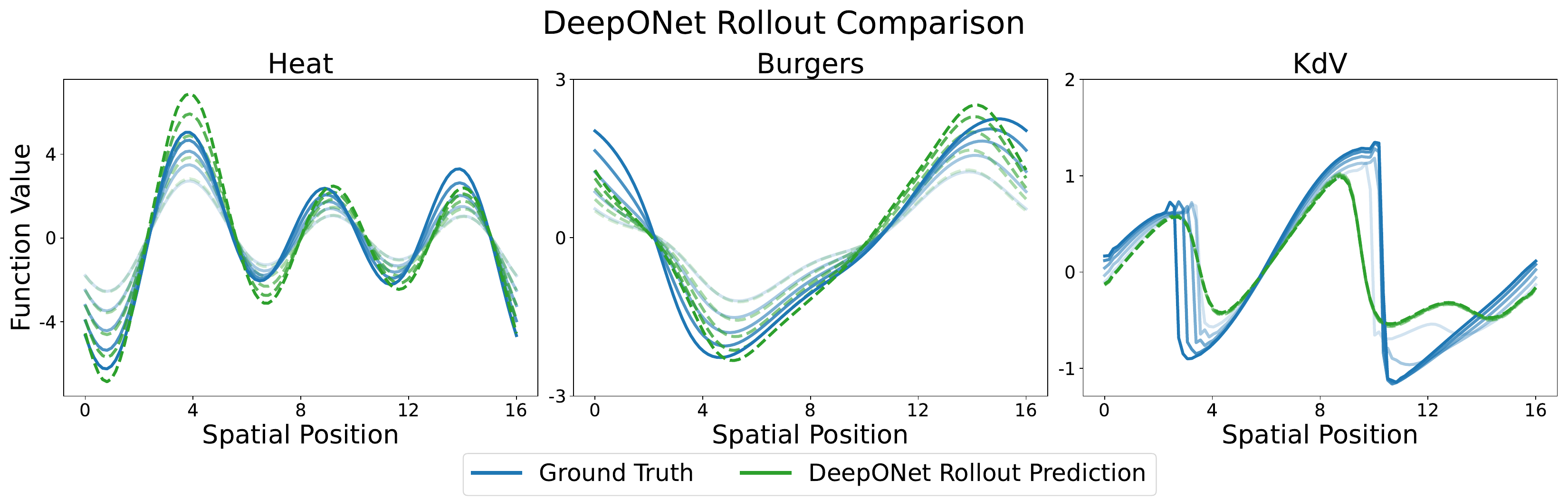}
    \caption{Comparison of DeepONet to ground truth data for autoregressive rollout on our 1D data sets.}
    \label{fig:pitt_1d_rollout}
\end{figure}
\begin{figure}[H]
    \centering
    \includegraphics[width=\linewidth]{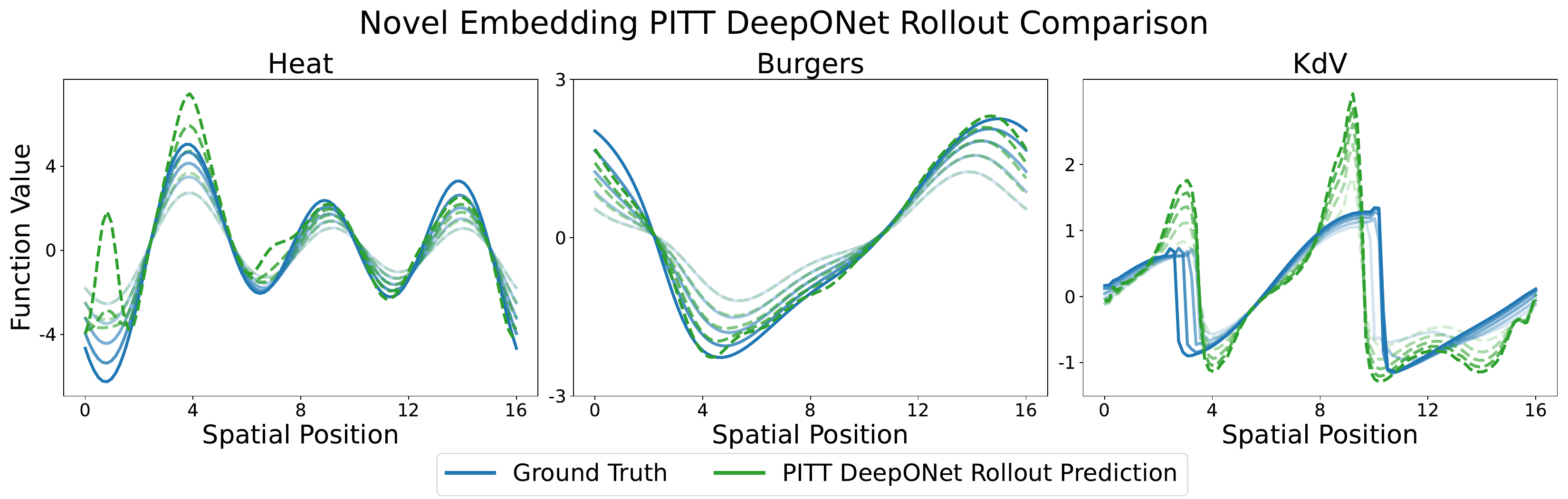}
    \caption{Comparison of PITT DeepONet using our novel embedding to ground truth data for autoregressive rollout on our 1D data sets.}
    \label{fig:pitt_2d_rollout}
\end{figure}
\begin{figure}[H]
    \centering
    \includegraphics[width=\linewidth]{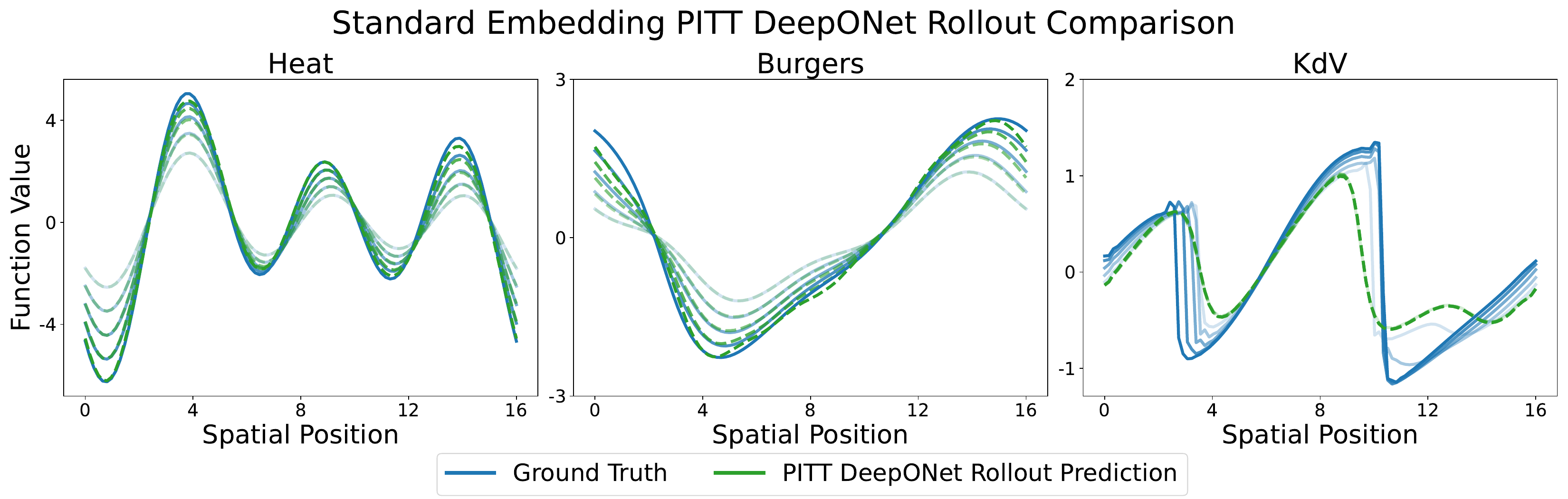}
    \caption{Comparison of PITT DeepONet using standard embedding to ground truth data for autoregressive rollout on our 1D data sets.}
    \label{fig:pitt_2d_rollout}
\end{figure}

\newpage
\section{2D Fixed Future Comparison}
In the 2D fixed-future experiments we see PITT is able to predict the finer detail better than all of the baseline models for both $T=20$ and $T=30$.
\label{app:2d_ff}
\begin{figure}
    \centering
    \includegraphics[width=\linewidth]{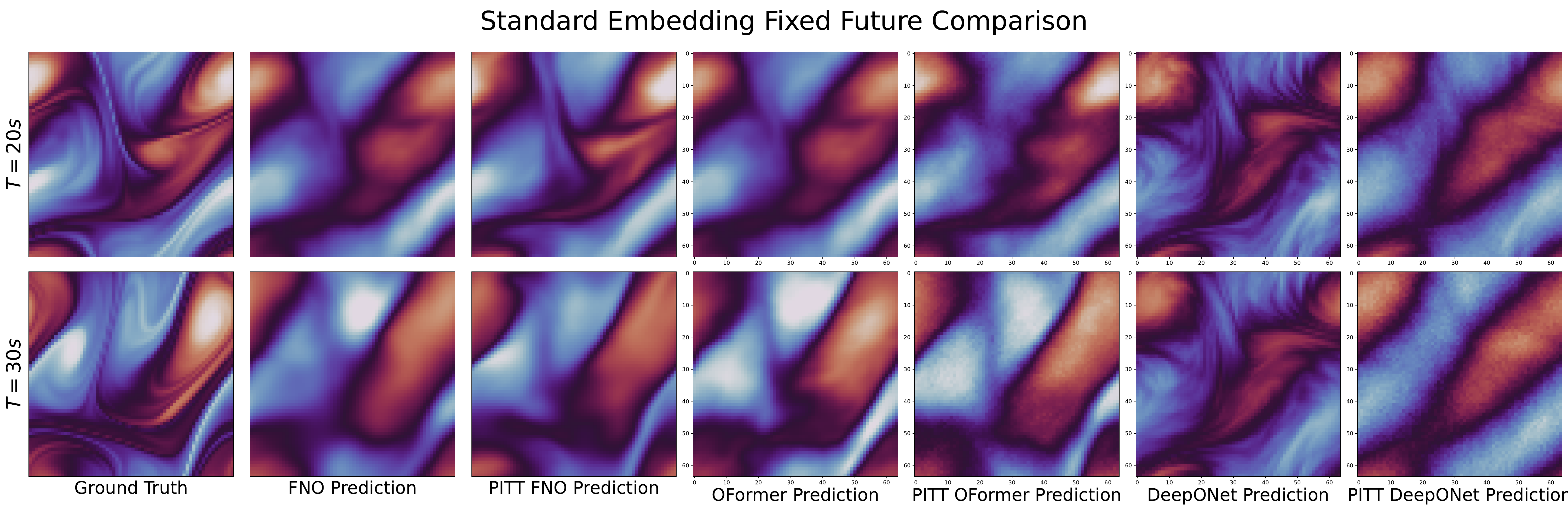}
    \caption{Comparison of fixed-future Navier-Stokes predictions between PITT variants and baseline models.}
    \label{fig:old_ff}
\end{figure}
\begin{figure}
    \centering
    \includegraphics[width=\linewidth]{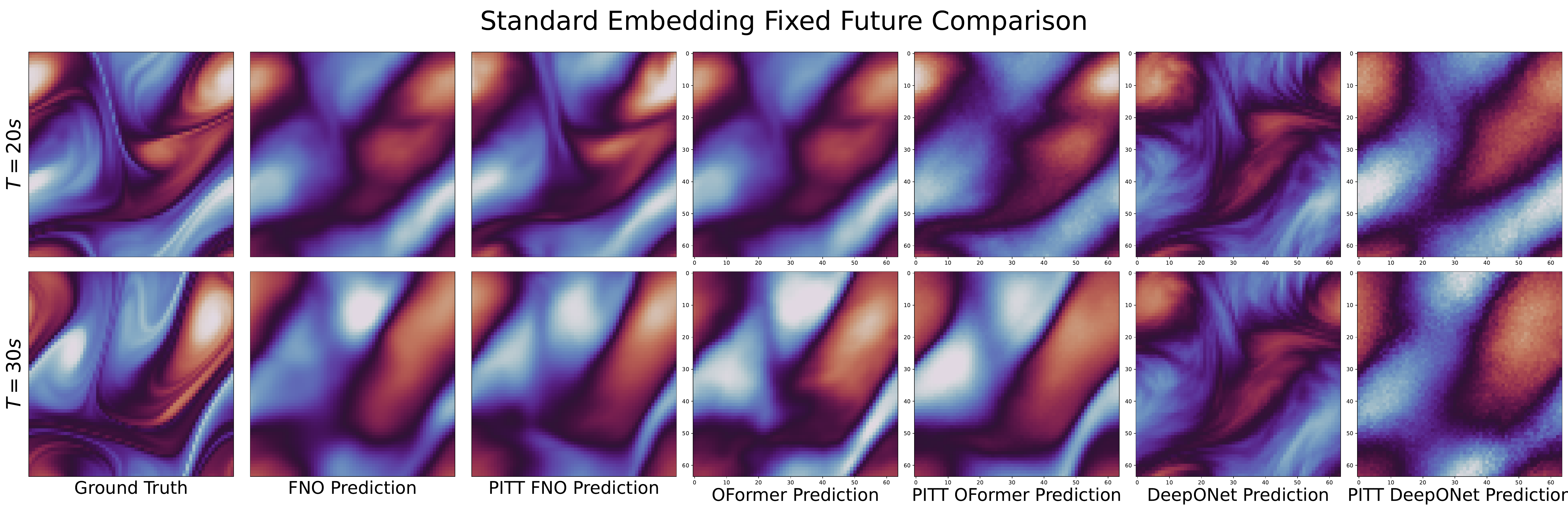}
    \caption{Comparison of fixed-future Navier-Stokes predictions between PITT variants and baseline models using our novel embedding.}
    \label{fig:new_ff}
\end{figure}

\newpage
\section{2D Poisson Comparison}
In the 2D Poisson equation, we see PITT has significantly less error across the entire prediction domain for all PITT variants when compared to the baseline model.
Here darker regions indicate higher error.
\label{app:2d_poisson}

\begin{figure}
    \centering
    \includegraphics[width=0.8\linewidth]{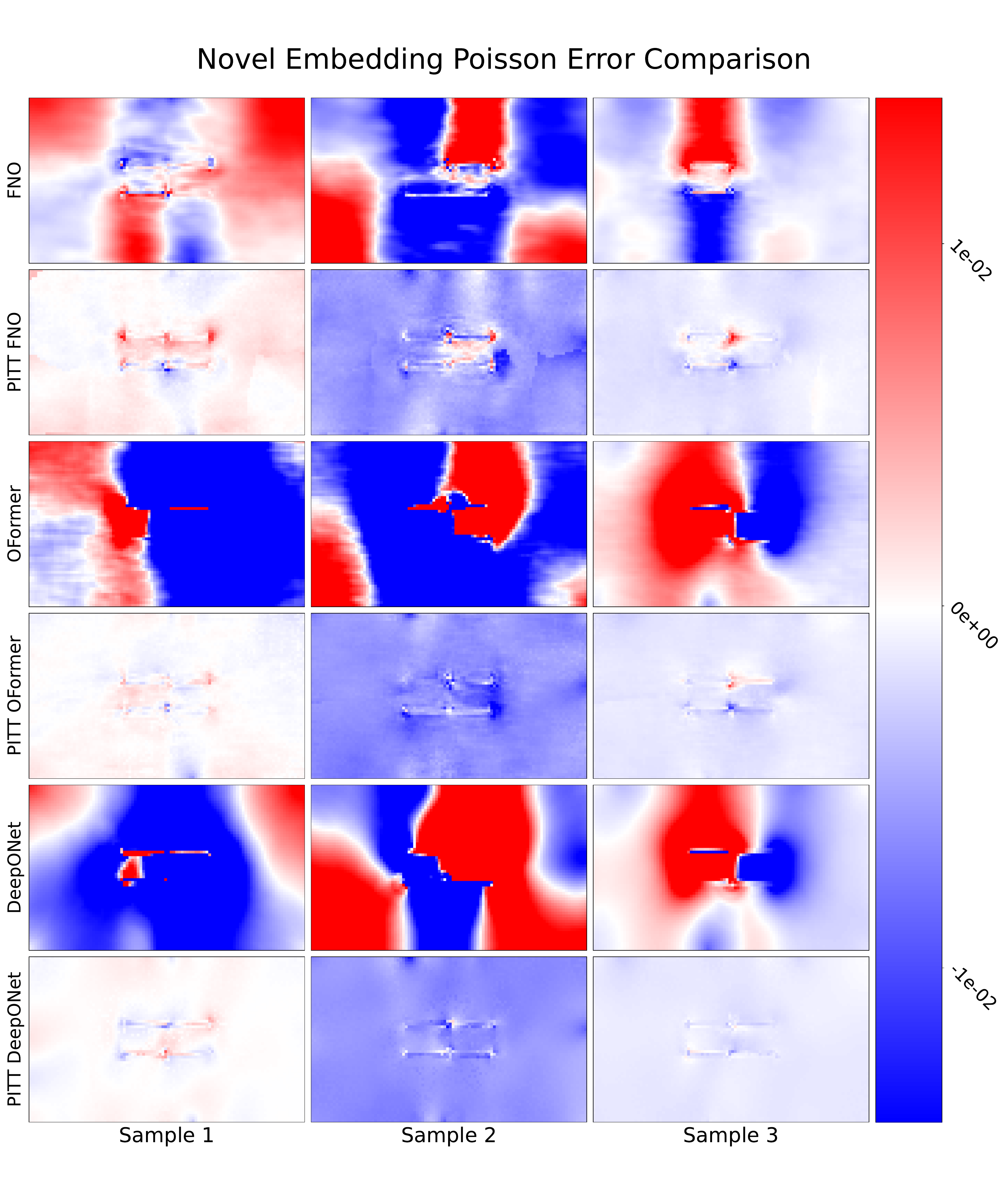}
    \caption{Comparison of Poisson prediction error between PITT variants and baseline models using novel embedding.}
    \label{fig:old_poisson_error_comp}
\end{figure}
\begin{figure}
    \centering
    \includegraphics[width=0.8\linewidth]{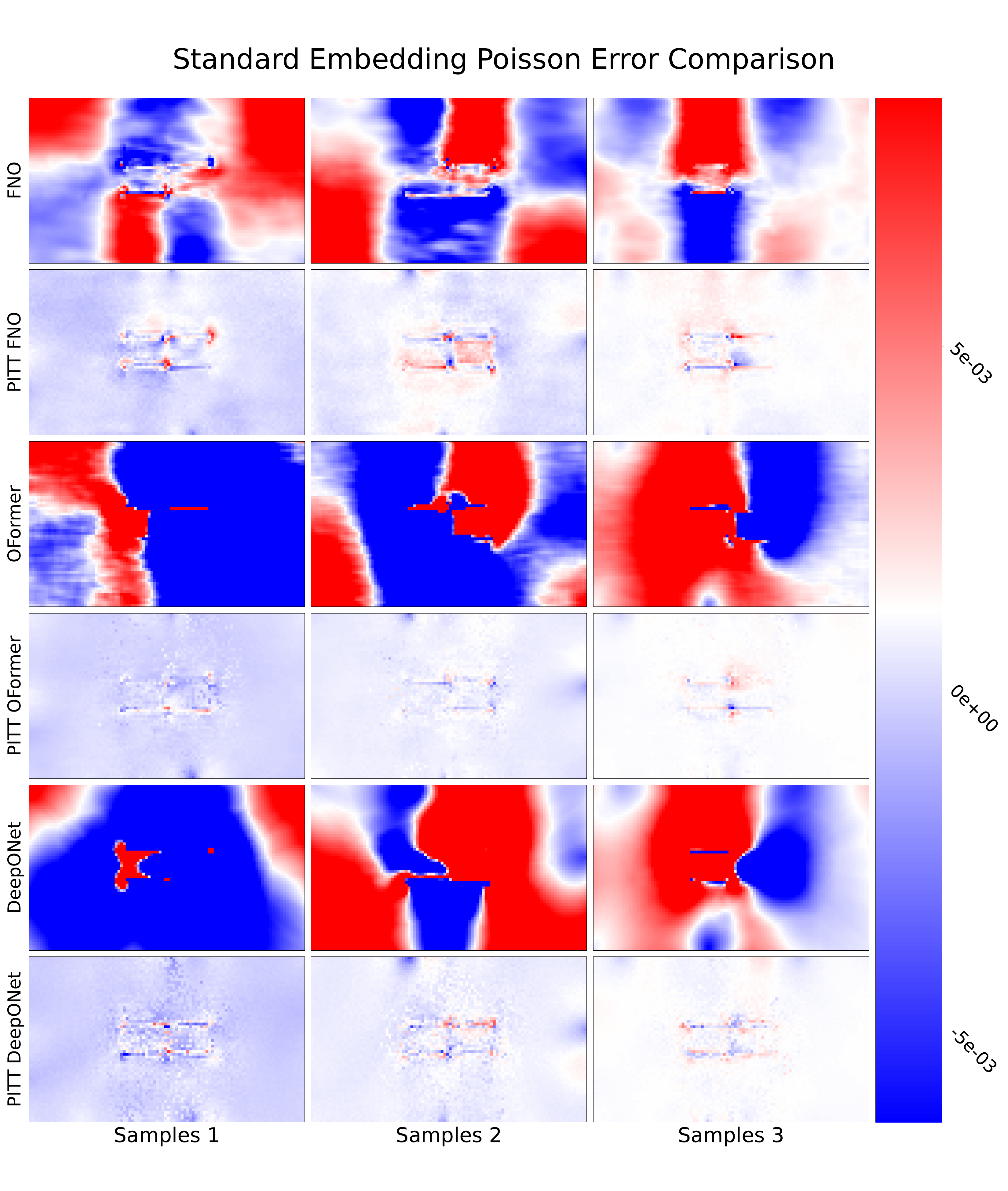}
    \caption{Comparison of Poisson predicition error between PITT variants and baseline models using standard embedding.}
    \label{fig:new_poisson_error_comp}
\end{figure}

\newpage
\section{2D Rollout Comparison}
\begin{figure}
    \centering
    \includegraphics[width=0.9\linewidth]{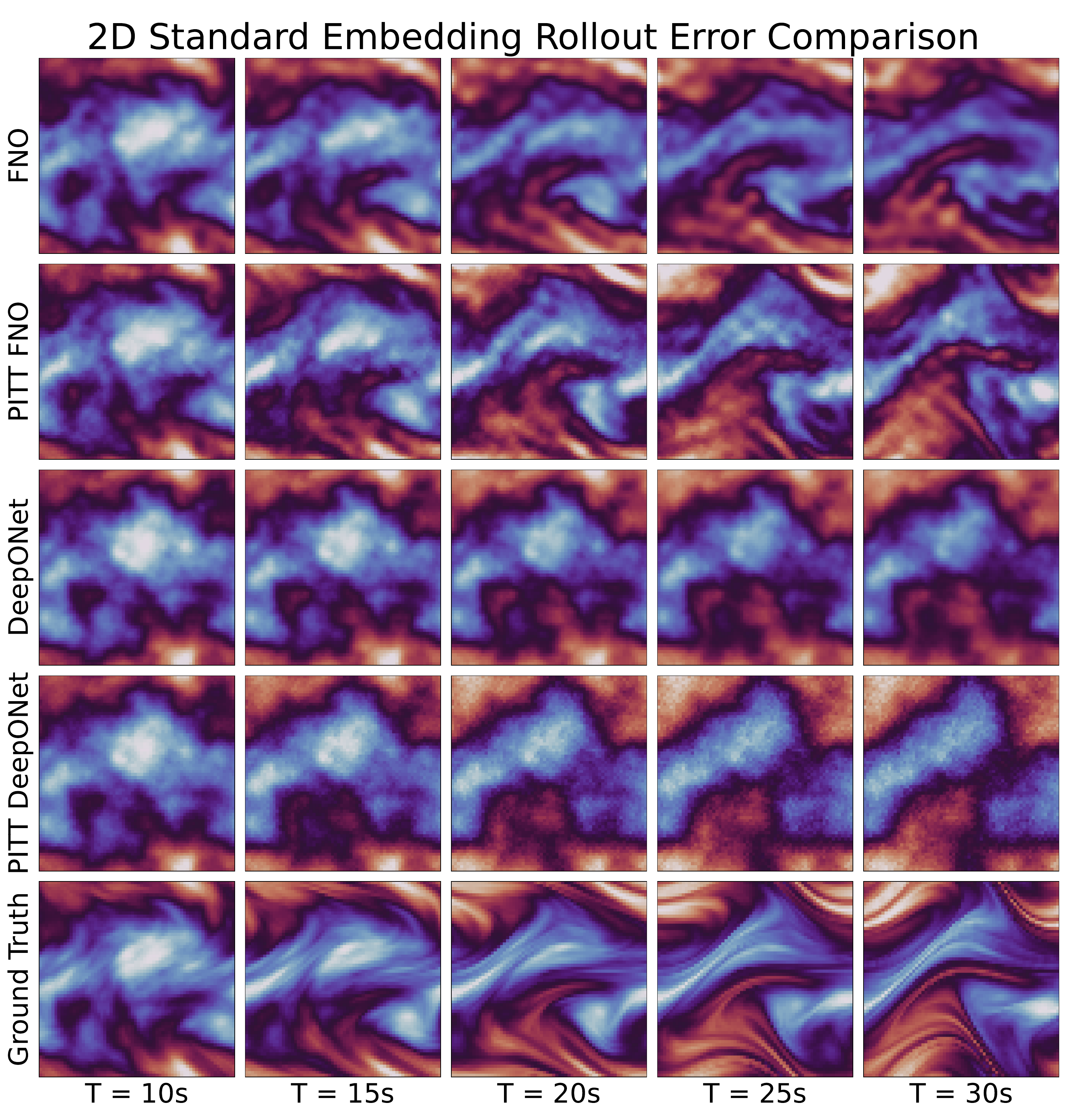}
    \caption{Rollout results for 2D Navier Stokes using standard embedding.}
    \label{fig:enter-label}
\end{figure}

\newpage
\section{Time-to-Solution}

In this section we roughly determine time-to-solution for our 1D experiments.
We see that the baseline models and their PITT variants are faster than numerical methods across all 1D data sets for both next-step prediction and fixed-future, final state prediction.
PITT DeepONet is faster than baseline DeepONet due to smaller linear layers: 256 for our baseline and 128 for our PITT variants, seen in tables \ref{tab:deeponet_ns_1d_model_hyperparams} and \ref{tab:deeponet_ff_1d_hyperparameters}.
The KdV equation has much longer time-to-solution than other equations due to the adaptive time step in our fourth-order Runge-Kutta Dormand–Prince solver.
The very fine details present in KdV solutions require much finer adaptive timesteps to resolve.
We calculate the average timestep and average number of adaptive refinement steps for each equation across all coefficient combinations.
For the Heat equation, we have an average timestep of 0.03584, requiring 0.25025 adaptive steps on average, for Burgers' equation, we have an average timestep of 0.03531, requiring 0.27783 adaptive steps on average, and for the KdV equation, we have an average timestep of 0.00032 requiring 7.33064 adaptive steps on average.
Additional computational overhead in KdV simulation is due to the recursive nature of our numerical solver, where finer temporal discretizations require multiple recursive steps to determine convergence before proceeding with the simulation step.
For our next-step case, the numerical time results were averaged over 100 timesteps, with 100 Heat equation samples for each $\beta$ value, for 600 total samples, $100$ Burgers equation samples for each $\alpha$ and $\beta$ combination, for 3600 total samples, and 5 KdV equation samples for each $\alpha$ and $\gamma$ combination, for 30 total samples.
In the fixed-future case, we average over the total simulation time for each coefficient combination of each equation.
For the next step timing results, each model was averaged over 90 steps from 200 samples for equation.
For the fixed-future experiment, each model was averaged over 200 samples for each equation.
Timing results were done using PyTorch 1.13.0 on a GeForce 2080 TI GPU, and our numerical simulations were done on an Intel(R) Core(TM) i9-9900K CPU @ 3.60GHz using PyTorch 1.10.2, since GPU runs took longer.

%
%

\begin{table}[H]
    \centering
    \resizebox{\linewidth}{!}{
    \begin{tabular}{c|c|ccccccccc}
         Data Set & Numerical & FNO & PITT FNO$^{\dagger}$ & PITT FNO$^{*}$ & OFormer & PITT OFormer$^{\dagger}$ & PITT OFormer$^{*}$ & DeepONet & PITT DeepONet$^{\dagger}$ & PITT DeepONet$^{*}$ \\
         \hline
         Heat & 0.00531 & 0.00162 & 0.00261 & 0.00274 & 0.00335 & 0.00456 & 0.00466 & 0.000281 & 8.681e-08 & 8.624e-08 \\
         Burgers & 0.00539 & 0.00138 & 0.00269 & 0.00271 & 0.00338 & 0.00464 & 0.00481 & 0.000294 & 8.860e-08 & 9.220e-08\\
         KdV & 1.238 & 0.00135 & 0.00272 & 0.00271 & 0.00333 & 0.00587 & 0.00472 & 0.000354 & 1.116e-07 & 8.769e-08 \\
    \end{tabular}}
    \caption{Next Step Prediction Time (s)}
    \label{tab:ns_pred_time}
\end{table}

\begin{table}[H]
    \centering
    \resizebox{\linewidth}{!}{
    \begin{tabular}{c|c|ccccccccc}
         Data Set & Numerical & FNO & PITT FNO$^{\dagger}$ & PITT FNO$^{*}$ & OFormer & PITT OFormer$^{\dagger}$ & PITT OFormer$^{*}$ & DeepONet & PITT DeepONet$^{\dagger}$ & PITT DeepONet$^{*}$ \\
         \hline
         Heat & 0.526 & 0.00540 & 0.00694 & 0.00350 & 0.00322 & 0.00619 & 0.00622 & 0.000286 & 1.138e-05 & 1.173e-05 \\
         Burgers & 0.533 & 0.00127 & 0.00337 & 0.00339 & 0.00325 & 0.00624 & 0.00624 & 0.000280 & 1.148e-05 & 1.159-05 \\
         KdV & 122.570 & 0.00127 & 0.00336 & 0.00338 & 0.00323 & 0.00619 & 0.00625 & 0.000284 & 1.146e-05 & 1.181e-05 \\
    \end{tabular}}
    \caption{Fixed-Future Prediction Time (s)}
    \label{tab:ff_pred_time}
\end{table}

\end{document}